\ifdefined\pdfoutput\pdfoutput=1\fi
\PassOptionsToPackage{numbers,sort&compress}{natbib}
\documentclass[11pt,a4paper,logo]{lumia}
\usepackage{natbib}

\usepackage{amsmath,amsfonts,bm}

\def\eqref#1{equation~\ref{#1}}

\def\1{\bm{1}}

\DeclareMathAlphabet{\mathsfit}{\encodingdefault}{\sfdefault}{m}{sl}
\SetMathAlphabet{\mathsfit}{bold}{\encodingdefault}{\sfdefault}{bx}{n}

\usepackage{wrapfig,xurl}
\tcbuselibrary{breakable}

\definecolor{abstrabg}{HTML}{F2D1C5}
\definecolor{abstradark}{HTML}{D97757}
\definecolor{reportaccent}{HTML}{BB0819}
\colorlet{reporttoc}{reportaccent}
\definecolor{cayenne}{HTML}{941100}
\hypersetup{allcolors=cayenne,
  pdfauthor={Hao Wang et al.},
  pdftitle={OPIS: An Input-Grounded Benchmark for Multi-Object Memory in Video World Models}}
\colorlet{OPISNavy}{reportaccent}
\colorlet{OPISTeal}{cayenne}
\colorlet{OPISLight}{abstrabg!35!white}
\definecolor{OPISGray}{HTML}{5D6873}
\colorlet{OPISBlue}{reportaccent}
\tcbset{colback=OPISLight,colframe=reportaccent,colbacktitle=reportaccent,coltitle=white}
\newcommand{\resourceicon}[1]{\makebox[1.5em][c]{\color{cayenne}#1}}
\setheadertext{OPIS}
\setheadertitle{OPIS: An Input-Grounded Benchmark for Multi-Object Memory in Video World Models}
\title{OPIS: An Input-Grounded Benchmark for\\
  Multi-Object Memory in Video World Models}
\makeatletter
\gdef\@author{\parbox{\textwidth}{\centering
{\Authfont
Hao Wang$^{1,4,*,\ddag}$,\ Tao Yu$^{2,*,\dagger}$,\ Liuzhou Zhang$^{3,*}$,\ HeXin Wang$^{4}$,\
Haopeng Jin$^{2}$,\ Yuxuan Zhou$^{5}$,\\[-0.05em]
Xinming Wang$^{2}$,\ Hongzhu Yi$^{6}$,\ Xinye Li$^{7}$,\ Yuanlei Wang$^{8}$,\
Ping Nie$^{9}$,\ Yan Huang$^{2,12}$,\\[-0.05em]
Yuxuan Zhang$^{10}$,\ Pengfei Zhou$^{4,11,\dagger}$,\ Yanyan Zou$^{13}$,\ Wei Yang$^{1,\dagger}$}
\\[0.55em]
\begin{minipage}{0.96\textwidth}
\centering\Affilfont
$^{1}$USTC \quad
$^{2}$CASIA \quad
$^{3}$HKUST \quad
$^{4}$Infrec \quad
$^{5}$Tsinghua University \quad
$^{6}$UCAS \quad
$^{7}$The Chinese University of Hong Kong\\
$^{8}$Sun Yat-sen University \quad
$^{9}$University of Waterloo \quad
$^{10}$Jiangnan University \quad
$^{11}$NUS \quad
$^{12}$Fiveages \quad
$^{13}$SUTD
\end{minipage}
\\[0.55em]
\begin{minipage}{0.96\textwidth}
\centering\Affilfont
$^{*}$Equal contribution. \qquad $^{\dagger}$Corresponding authors.  \qquad $^{\ddag}$Work done during an internship at Infrec Tech.
\end{minipage}
}}
\makeatother
\date{September 26, 2026}
\newcommand{\reportdate}{2026-09-26}
\newcommand{\paperresources}{%

\linkfont
\resourceicon{\faCalendar}\textbf{Date:}\enspace September 26, 2026\\[0.2em]
\resourceicon{\faGithub}\textbf{Code:}\enspace \href{https://github.com/SSStarain/OPIS}{https://github.com/SSStarain/OPIS}\\[0.2em]
\resourceicon{\faDatabase}\textbf{Data:}\enspace \href{https://huggingface.co/datasets/Kirito-Lab/OPIS-dataset}{https://huggingface.co/datasets/Kirito-Lab/OPIS-dataset}

\par}
\makeatletter
\renewcommand{\lumia@githublink}{\paperresources}
\makeatother

\newcommand{\Obj}{\mathcal{O}}
\newcommand{\RefSet}{\mathcal{R}}

\newcommand{\Score}{\mathrm{Score}}

\newcommand{\cmark}{\ding{51}}
\newcommand{\xmark}{\ding{55}}
\newcommand{\pmark}{\ensuremath{\triangle}}

\newcommand{\tablecaption}[1]{%
  \setlength{\abovecaptionskip}{0pt}%
  \setlength{\belowcaptionskip}{0pt}%
  \vspace{\baselineskip}%
  \caption{#1}%
  \vspace{\baselineskip}%
}

\begin{document}

\begin{abstract}
Video world models must preserve the visual state of the world over time, but existing evaluation protocols often rely on generated histories, video reference, or selected revisit viewpoints that can confound the assessment of a model's true memory capability. To address this, we introduce OPIS, an input-grounded benchmark that strictly anchors the assessment to a fixed set of object instances from the initial observation for evaluating multi-object memory in video world models. The OPIS dataset comprises 500 cases across real-world, embodied-robotic, and game-world domains, providing dense object-level annotations for 12,672 rigid, articulated, and deformable instances. Our object-centric evaluator combines association and explicit visibility reasoning to hierarchically measure Object (O) Presence (P), Identity (I), and Structure (S), utilizing static or dynamic evaluation tracks based on object kinematics. Across eight image-to-video or camera-conditioned world models, our proposed OPIS scores range from 48.65 to 56.01. As the reference inventory grows from less than 20 to more than 40 objects, the Presence, Identity, and Structure scores show an overall decline, with the average Identity score falling from 40.22 to 23.11. The results demonstrate that preserving the particular object instances in the input is considerably harder than generating plausible visual elements.
\end{abstract}

\teaserfigure{
\centering
\includegraphics[width=\textwidth]{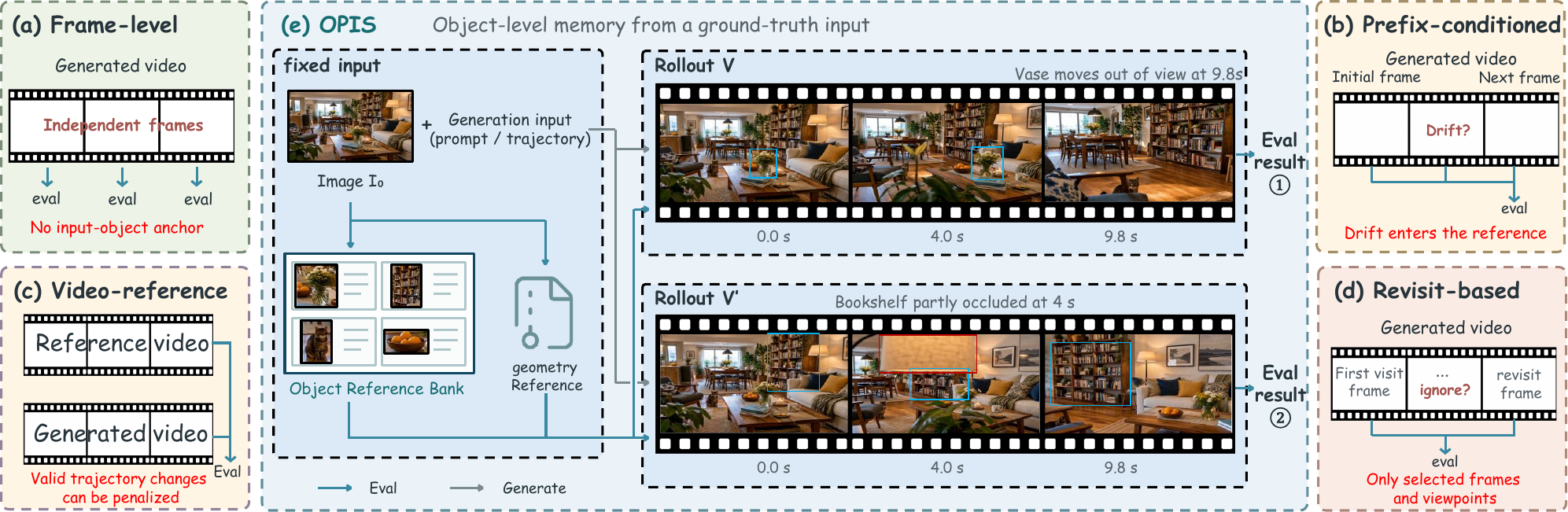}
\captionof{figure}{Comparison of evaluation paradigms and the input-grounded approach. Panels (a)--(d) illustrate evaluation using independent frames, generated prefixes, reference videos, and selected revisits. Panel (e) anchors both rollouts to the same object reference bank and geometry.}
\label{fig:opis-overview}
}

\maketitle
\clearpage
\begingroup
\hypersetup{linkcolor=reporttoc,linktoc=section}
\setlength{\parskip}{0pt}
\setcounter{tocdepth}{3}
\tableofcontents
\endgroup
\clearpage

\section{Introduction}
\label{sec:intro}
Video world models are increasingly being used as visual simulators, with systems spanning learned latent dynamics~\citep{Ha2018WorldModels,Hafner2023DreamerV3}, interactive environment generation~\citep{Bruce2024Genie}, video prediction for autonomous driving~\citep{Hu2023GAIA1}, and long-term spatial memory~\citep{Wu2025LongTermSpatialMemory}. Given an initial observation and an instruction, they must carry the observed world forward over time while preserving the state that makes it the same world. For these models, the initial observation is the only source of ground-truth visual information. It is therefore essential to distinguish the input, the state carried forward from it, and the generated output.

Existing protocols offer complementary views of generated worlds through independent-frame quality~\citep{Huang2024VBench}, comparison between complete generated and reference videos~\citep{MIND2026}, consistency with generated history~\citep{WorldTrace2026}, and recovery at revisits~\citep{MBench2026,R2MBench2026} (Figure~\ref{fig:opis-overview}, a--d). Although valuable for their intended tasks, these protocols do not fully isolate the preservation of the world observed in the input. Independent-frame evaluation lacks an input-object anchor; video-reference comparison may penalize valid trajectory differences; generated-history comparison can inherit accumulated drift; and revisit tests cover only selected frames and viewpoints. Meanwhile, image-level scores may conceal missing objects, identity changes, or structural degradation within otherwise plausible scenes. These limitations motivate evaluating individual objects throughout generation against a fixed reference derived solely from the initial observation, with explicit reasoning about their visibility (Figure~\ref{fig:opis-overview}, e). The same input reference can support different valid rollouts, while visibility reasoning distinguishes memory failures from events such as a vase moving out of view or a bookshelf becoming partially occluded.

We introduce \textbf{OPIS}, an input-grounded benchmark for multi-object memory in video world generation. Our annotation pipeline constructs dense object-level references with instance masks, appearance descriptors, and mobility labels, together with an auxiliary, precomputed reference geometric representation. The dataset contains 500 cases across 3 scene domains (real worlds, embodied-robotic, and game worlds) and 10 subcategories. It covers 12,672 object instances: 8,432 rigid, 2,046 articulated, and 2,194 deformable objects. The evaluator is deliberately evidence-first. It compares sampled generated frames against the fixed input reference through one-to-one association and explicit visibility reasoning, measuring \emph{Presence}, whether instances are accounted for when observable; \emph{Identity}, whether observed instances retain their input identities; and \emph{Structure}, whether their geometry or structural properties persist. Based on object kinematics and motion expectations, Structure uses two tracks: static geometry for rigid objects expected to remain stationary, and dynamic structure for the remaining objects, assessing input-grounded structural properties while allowing motion, articulation, and deformation.

Experiments on eight image-to-video (i2v) and camera-conditioned world models show that object accounting is substantially easier than preserving the particular input instances. Presence remains acceptable across models (85.63--93.49), whereas Identity is much lower (28.12--36.88). Seedance 2.0~\citep{seedance2026seedance20advancingvideo}, one of the leading i2v models, achieves the highest overall OPIS score (56.01), while Echo-WM-Flash~\citep{zhang2026echowmopenenterableomnimodal}, a recently released camera-conditioned world model, leads the evaluated camera-conditioned group (51.42). The difficulty scales with the number of addressable objects: average Identity falls from 40.22 for scenes with at most 20 reference object instances to 23.11 for scenes with more than 40. These findings show that input-grounded, object-centric evaluation exposes memory failures that aggregate video-quality measures can conceal.

Our contributions are fourfold:
\begin{itemize}
      \item \textbf{An input-grounded formulation of multi-object memory.} We anchor evaluation to a fixed set of object instances from the initial observation, measuring their persistence under camera motion and partial visibility without requiring a reference continuation.
      \item \textbf{A cross-domain dataset with fine-grained annotations.} OPIS comprises 500 cases across 3 scene domains and 10 subcategories, covering 12,672 rigid, articulated, and deformable instances. It provides dense object-level annotations.
      \item \textbf{An object-centric, hierarchical evaluator.} We assess object memory through Presence, Identity, and Structure, combining object-level association and explicit visibility reasoning with static or dynamic evaluation tracks selected according to each object's kinematics.
      \item Experiments across eight models reveal a persistent gap between object presence and object identity and show that object memory preservation declines as the number of addressable objects increases.
\end{itemize}

\section{Related Work}
\label{sec:related}

\subsection{Evaluation References and Memory}

Table~\ref{tab:related-eval} compares evaluation references and object-level evidence. Frame-quality components in VBench~\citep{Huang2024VBench} assess individual images without testing input-instance preservation. Reference-video tests in MIND~\citep{MIND2026} and Latent Spatial Memory (LSM)~\citep{LatentSpatialMemory2026} measure reconstruction under prescribed trajectories; their errors can also reflect alternative valid continuations in open-ended generation. Local consistency in WorldScore~\citep{duan2025worldscoreunifiedevaluationbenchmark} and WorldTrace~\citep{WorldTrace2026} uses generated history, which may already contain drift. Revisit tests in MBench~\citep{MBench2026}, LoopBench~\citep{WorldTrace2026}, and R2M-Bench~\citep{R2MBench2026} probe selected return views; R2M additionally controls for generic temporal stability.

\Needspace{22\baselineskip}
\begin{wraptable}{r}{0.50\textwidth}
\centering
\scriptsize
\setlength{\tabcolsep}{1.5pt}
\renewcommand{\arraystretch}{1.08}
\caption{Evaluation design comparison.}
\label{tab:related-eval}
\begin{tabular*}{\linewidth}{@{\extracolsep{\fill}}cccccc@{}}
\toprule
Method & \shortstack{Eval.\\protocol} & \shortstack{Object-\\level}
& \shortstack{Fixed\\{}input set} & \shortstack{Visib.-\\aware}
& \shortstack{Object\\struct.} \\
\midrule
VBench & F, PW & \pmark & \xmark & \xmark & \xmark \\
VBench-2.0 & F, PW & \pmark & \xmark & \pmark & \pmark \\
MIND & F, VR, R & \xmark & \xmark & \xmark & \xmark \\
MBench & PW, R & \pmark & \xmark & \pmark & \pmark \\
R2M-Bench & R, PW & \pmark & \xmark & \pmark & \pmark \\
WorldTrace & PW, R & \xmark & \xmark & \xmark & \xmark \\
LSM & VR, R, IG & \pmark & \xmark & \xmark & \xmark \\
WorldScore & F, PW & \pmark & \xmark & \xmark & \xmark \\
WBENCH & F, PW, R & \pmark & \pmark & \pmark & \pmark \\
T2V-CompBench & F & \pmark & \xmark & \xmark & \xmark \\
PDI-Bench & PW & \cmark & \xmark & \pmark & \cmark \\
\midrule
\textbf{OPIS (ours)} & \textbf{IG} & \cmark & \cmark & \cmark & \cmark \\
\bottomrule
\end{tabular*}
\par\smallskip
\raggedright
Symbols denote explicit
(\cmark), partial (\pmark), or no explicit (\xmark) support under
the definitions below. F: frame-level; VR: video-reference; PW: generated-prefix/window; R: revisit; IG: input-grounded.
\end{wraptable}
The initial-frame return comparison in LSM~\citep{LatentSpatialMemory2026} and the first-frame subject anchor in WBENCH~\citep{ying2026wbenchcomprehensivemultiturnbenchmark} show that input or first-frame anchoring already has precedents.

\subsection{Object-Level Evidence}
Object-level evaluation is also established: T2V-CompBench~\citep{T2VCompBench2025} assesses text-grounded composition, VBench-2.0~\citep{Zheng2025VBench20} includes human identity and anatomy, and MBench~\citep{MBench2026} and R2M-Bench~\citep{R2MBench2026} measure entity consistency. Visibility handling ranges from valid-observation filtering~\citep{ying2026wbenchcomprehensivemultiturnbenchmark} to occlusion-aware VLM judgments~\citep{R2MBench2026}; PDI-Bench~\citep{PDI2026} directly measures object rigidity from reconstructed trajectories. These mechanisms differ from retaining a fixed input inventory and distinguishing missing instances from unobservable ones. OPIS combines that inventory with one-to-one association, explicit visibility and unknown states, and object-level Presence, Identity, and Structure. Its distinction is this integrated evaluation contract, rather than object scoring or first-frame anchoring alone.

\section{OPIS Dataset}
\label{sec:dataset}

\subsection{Task and Scope}
Each OPIS case consists of an initial image $I_0$, a generation instruction $u$, and a fixed evaluator-side reference $\RefSet_0$:
\begin{equation}
    d=(I_0,u,\RefSet_0),\qquad
    V=G_\theta(I_0,u),\qquad
    \RefSet_0=(\Obj_0,\mathcal{G}_0).
    \label{eq:case-contract}
\end{equation}
The instance set $\Obj_0$ records the objects visible in the input, while $\mathcal{G}_0$ provides auxiliary geometric evidence derived from the same image. The generator receives only $(I_0,u)$; the reference is constructed once before generation and is never updated with generated content. OPIS therefore evaluates how well a rollout preserves input-grounded object evidence across time, without requiring a target continuation or access to the model's internal memory.

The instructions are designed to expose memory under changing viewpoints, motion, and partial visibility. They encourage smooth camera movement, parallax, and re-observation of selected objects, including cases in which an object leaves view and later reappears. These instructions define the evaluation challenge rather than a prescribed future trajectory. OPIS scores observable memory while allowing valid changes in camera pose, object motion, articulation, and deformation.

\subsection{Data Coverage and Sources}

\begin{figure*}[t]
\centering
\includegraphics[width=\textwidth]{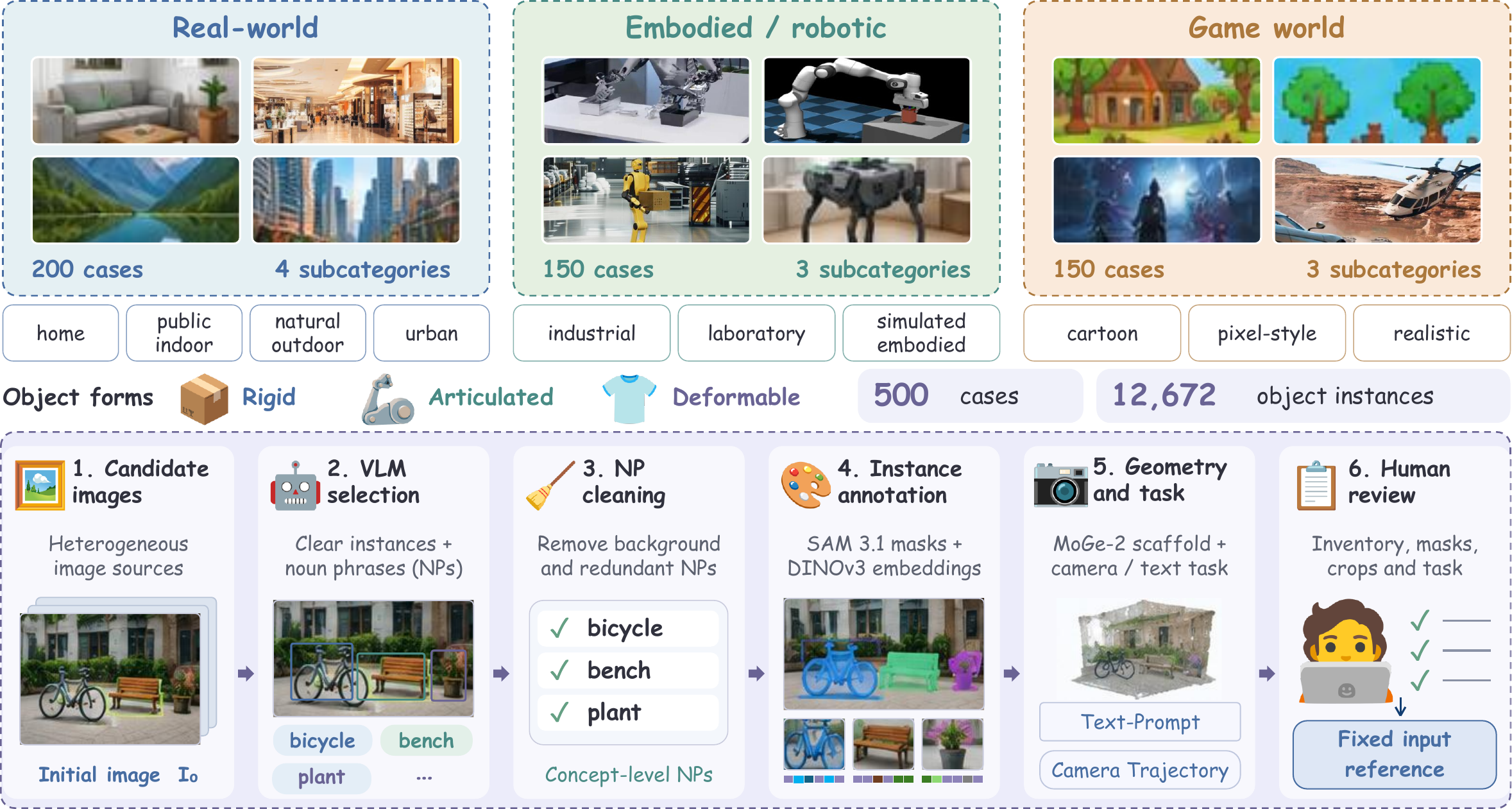}
\caption{OPIS dataset composition and construction pipeline. Top: 500 cases span real-world, embodied-robotic, and game-world domains, with ten subcategories and rigid, articulated, and deformable objects. Bottom: candidate images undergo VLM-assisted selection, noun-phrase (NP) cleaning, instance annotation, geometry and task construction, and human review to produce a fixed input reference.}
\label{fig:evaluation-pipeline}
\end{figure*}

The OPIS dataset contains 500 cases spanning three scene domains and ten subcategories (Figure~\ref{fig:evaluation-pipeline}). Each subcategory contains 50 cases: home, public indoor, natural outdoor, and urban scenes; industrial, laboratory, and simulated embodied environments; and cartoon, pixel-style, and realistic game worlds. Across the benchmark, 12,672 addressable rigid, articulated, and deformable instances support memory evaluation across varied object densities and motion expectations. Domain-level counts and evaluation-track assignments are detailed in Appendix~\ref{app:dataset-composition} (Table~\ref{tab:dataset-domains}).

OPIS combines real-world imagery, game screenshots, simulator-rendered scenes, and images generated with gpt-image-2.5-sunburst. Real-world sources include Visual Genome~\citep{Krishna2017VisualGenome}, ADE20K~\citep{Zhou2017ADE20K}, COCO~\citep{Lin2014COCO} images indexed through RefCOCO~\citep{Nagaraja2016ReferringExpressions}, and embodied or industrial collections such as BridgeData V2~\citep{Walke2023BridgeData} and IndustryShapes~\citep{sapoutzoglou2026industryshapesrgbdbenchmarkdataset}. Game imagery comes from gameplay-caption collections, including Minecraft~\citep{Fan2022MineDojo} and SuperTuxKart~\footnote{https://supertuxkart.net/} scenes; locally rendered AI2-THOR~\citep{Kolve2017AI2THOR} images provide simulated embodied environments. Together with the generated images, these sources provide complementary scene layouts, visual styles, and object configurations.

\subsection{Data Construction}
Our construction pipeline turns heterogeneous source images into a common input-grounded reference, combining VLM-assisted selection and annotation, concept-guided instance segmentation, and final human review (Figure~\ref{fig:evaluation-pipeline}).

\textbf{Image selection and object vocabulary.}
A VLM first screens candidate images for a sufficient number of distinguishable object instances with clear boundaries, while proposing a list of noun phrases (NPs) describing the visible objects. When source annotations or simulator metadata already provide suitable NPs, we prioritize that vocabulary. Each NP denotes a concept and may correspond to multiple instances. A subsequent VLM-assisted cleaning stage removes background NPs and incomplete object phrases, and consolidates semantically similar or subsuming NPs.

\textbf{Instance segmentation and annotation.}
Using the cleaned NP list, we perform multiple rounds of perceptual concept segmentation (PCS) with SAM~3.1~\citep{Carion2025SAM3} to recover individual object instances. We retain each instance mask and its corresponding image crop, and compare masks using intersection-over-union (IoU) to remove highly overlapping duplicates across queries and rounds. DINOv3~\citep{Simeoni2025DINOv3} extracts an appearance embedding from each instance crop to support identity matching. Each instance is further assigned a persistent identifier, distinguishing its kinematic form. MoGe-2~\citep{Wang2025MoGeTwo} provides a fixed scene-level geometric scaffold from the initial image for viewpoint and occlusion reasoning.

\textbf{Instruction design and review.}
For each case, we construct a structured task specification including a camera trajectory and a text prompt. These plans define the intended memory challenge while allowing variation in the generated trajectory. A VLM assists with the initial specification, which is then individually reviewed together with the object inventory, masks, and crops to verify annotation quality and task coherence. Finally, each case is verified by humans.

\section{OPIS Evaluation}
\label{sec:pipeline}

\begin{wrapfigure}{r}{0.5\linewidth}
\centering
\includegraphics[width=\linewidth]{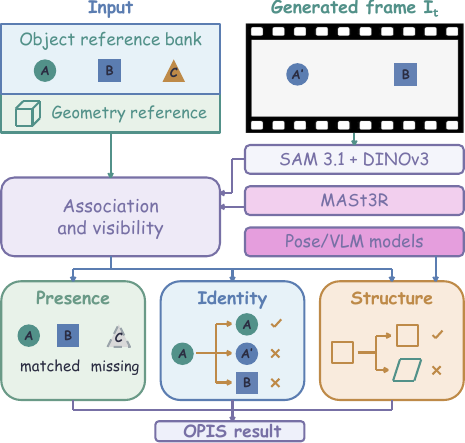}
\caption{Input-grounded evaluation of object presence, identity, and structure. Generated observations are associated with the fixed input reference using appearance and geometric evidence. The examples distinguish a missing object ($C$, when expected visible), an appearance change ($A$ to $A'$), and a structural change; valid viewpoint and motion changes are allowed.}
\label{fig:qualitative-placeholder}
\end{wrapfigure}

OPIS evaluates memory through the persistence of individual objects, using the initial observation as a fixed reference throughout generation. The evaluator first establishes which instances can be associated and observed, then measures their \emph{Presence}, \emph{Identity}, and \emph{Structure} (Figure~\ref{fig:qualitative-placeholder}). This separates object accounting from appearance fidelity and structural preservation, while retaining uncertainty when the available evidence is insufficient. Although OPIS evaluates individual object instances, we assess frames and videos by the number rather than the proportion of failed instances, since quality within a fixed image area should not depend on object density.

\subsection{Association and Visibility}
For each sampled frame $I_t$, we independently extract object observations using concept-guided segmentation and appearance encoding. Each observation contains a mask, bounding box, category, appearance attributes, and embedding. We associate these observations with the fixed input inventory $\Obj_0$ through a partial one-to-one assignment constrained by the reference geometric representation, following bipartite matching formulations~\citep{Kuhn1955Hungarian} used in object detection and tracking~\citep{Carion2020DETR}. Pairwise affinity combines embedding, attribute, and category similarity, with weights renormalized over available evidence. Null assignments allow unmatched instances, and minimum appearance-evidence requirements prevent category agreement alone from establishing identity. The one-to-one constraint prevents a detected object from being matched to multiple reference instances, while small affinity gaps between competing candidates flag ambiguous matches. Visibility reasoning distinguishes an absent object from one that cannot be assessed. For static instances, we project the input geometry into the current frame and test image bounds, projected size, and depth ordering. Camera motion is estimated with MASt3R~\citep{Leroy2024MASt3R} correspondences supported by the non-target background, with all annotated target masks excluded from camera fitting. Held-out background matches and aligned MoGe-2 depth provide reliability checks. A matched observation can also establish visibility directly. Independently moving objects are assessed through direct observations, since a static reference projection cannot determine their current location. Confirmed occlusion and out-of-view states are excluded from presence accounting. 

\subsection{Presence and Identity}
OPIS uses a frame-level criterion: a confirmed missing or extra object invalidates Presence for the frame, while a failure on any evaluable instance invalidates Identity or Structure. This prevents well-preserved objects from concealing a single object-memory failure or an additional generated object. 

\textbf{Presence (P).}
Let $\mathcal{E}_t$ contain the input instances with evidence that they should be visible in frame $t$, and let $p_{i,t}\in\{0,1\}$ indicate whether instance $i$ is matched to an observation. Let $\mathcal{X}_t$ denote the confirmed unmatched observations after one-to-one association, with unresolved extra-object attribution excluded. For a frame with $\mathcal{E}_t\cup\mathcal{X}_t\ne\varnothing$,
\begin{equation}
    P_t=\mathbb{1}\!\left[\forall i\in\mathcal{E}_t,\ p_{i,t}=1\right]
        \mathbb{1}\!\left[|\mathcal{X}_t|=0\right].
    \label{eq:presence-score}
\end{equation}
Any confirmed missing instance or confirmed unmatched extra observation makes the frame score zero. Confirmed occluded, out-of-view, or too-small instances are excluded, while unresolved visibility or unattributable extras remain unknown. Presence therefore requires complete one-to-one accounting among the instances and observations supported by sufficient evidence.

\textbf{Identity (I).}
For an observed match, appearance fidelity is the nonnegative cosine similarity between its input and generated DINOv3 embeddings, $a_{i,t}=\max(0,\cos(e_i,\widehat e_{i,t}))$. Let $\mathcal{A}_t$ contain matches with valid appearance evidence. For $\mathcal{A}_t\ne\varnothing$,
let $\tau_I$ denote the identity acceptance threshold, and define
\begin{equation}
    I_t=\mathbb{1}\!\left[\min_{i\in\mathcal{A}_t}a_{i,t}\geq\tau_I\right]
        \frac{1}{|\mathcal{A}_t|}\sum_{i\in\mathcal{A}_t}a_{i,t}.
    \label{eq:identity-score}
\end{equation}
If any valid identity score falls below $\tau_I$, the entire frame receives zero; otherwise, it retains the mean fidelity of its valid matches. Missing objects contribute to P and do not supply an identity measurement.

\subsection{Structure Across Static and Dynamic Objects}
Structure evaluates properties that should persist despite valid changes in viewpoint or motion. Mobility annotations route rigid objects expected to remain stationary to a \emph{static-geometry} track; articulated, deformable, and other dynamic-track instances are assessed through \emph{dynamic structure}. The tracks share an input-grounded reference but use evidence appropriate to each object's physical characteristics.

\textbf{Static geometry.}
For a matched static instance, the estimated camera transform projects its fixed input geometry into the current frame. We measure the discrepancy between these projections and image correspondences within the reference and observed bounding boxes:
\begin{equation}
    \rho_{i,t}=\frac{\operatorname{median}_{n\in\mathcal{M}_{i,t}}
      \lVert\Pi(K_t(R_tX_n^0+T_t))-x_{n,t}\rVert_2}
      {\operatorname{diag}(b_{i,t})},
    \qquad s^{\mathrm{stat}}_{i,t}=\exp(-\rho_{i,t}/\lambda_g).
    \label{eq:static-structure}
\end{equation}
Here $\mathcal{M}_{i,t}$ contains instance-associated correspondences, $X_n^0$ is their input-derived geometry, and $b_{i,t}$ is the observed bounding box. Residuals use all selected correspondences rather than only camera-fit inliers. Reliable camera and visibility evidence and sufficient correspondences are required for scoring. The resulting measure captures projective preservation of the visible input structure.

\textbf{Dynamic structure.}
The hybrid evaluator first uses specialist pose models, ViTPose~\citep{Xu2022ViTPose} and ViTPose++~\citep{Xu2024ViTPosePlusPlus}, for supported articulated categories. Visible landmarks are lifted using monocular geometry, and corresponding segment lengths are compared after removing a single scale factor per instance:
\begin{equation}
    r_{i,t,e}=\log\frac{\ell_{i,t,e}}{\ell_{i,0,e}},\qquad
    s^{\mathrm{pose}}_{i,t}=\frac{1}{|\mathcal{E}_{i,t}|}
      \sum_{e\in\mathcal{E}_{i,t}}
      \exp\!\left(-\frac{|r_{i,t,e}-\operatorname{median}_{e'\in\mathcal{E}_{i,t}}r_{i,t,e'}|}{\tau}\right).
    \label{eq:dynamic-structure}
\end{equation}
This assesses segment proportions without directly penalizing joint angles or global pose; uniform scale changes are removed by normalization. For other objects, a VLM assesses localized structural claims established from the input image. Claims concern visible parts, connectivity, local shape, or material continuity, as appropriate to the object's kinematic class. Each judgment is \emph{supported}, \emph{contradicted}, or \emph{unknown}. The score averages the supported fraction within each evaluable claim family, then across families.

\subsection{Aggregation and Evidence Coverage}
Let $\mathcal{B}_t$ contain instances with valid structural measurements in frame $t$, using the static or dynamic track assigned to each instance. The structural score uses the per-instance structure acceptance threshold $\tau_S$:
\begin{equation}
    S_t=\mathbb{1}\!\left[\min_{i\in\mathcal{B}_t}s_{i,t}\geq\tau_S\right]
        \frac{1}{|\mathcal{B}_t|}\sum_{i\in\mathcal{B}_t}s_{i,t},
    \qquad \mathcal{B}_t\ne\varnothing.
    \label{eq:strict-structure}
\end{equation}
Thus, any valid structure score below $\tau_S$ invalidates the frame; otherwise, the frame retains the mean score over its valid instances. Let $\mathcal{F}$ be all sampled frames, let $\mathcal{F}_P$ contain frames with expected-visible reference evidence or confirmed extra-observation evidence, and let $\mathcal{F}_I$ and $\mathcal{F}_S$ contain frames with valid appearance and structural measurements, respectively. Case scores are
\begin{equation}
    P=\frac{\sum_{t\in\mathcal{F}_P}P_t}{|\mathcal{F}_P|},\quad
    I=\frac{\sum_{t\in\mathcal{F}_I}I_t}{|\mathcal{F}_I|},\quad
    S=\underbrace{\frac{|\mathcal{F}_S|}{|\mathcal{F}|}}_{C_S}
      \frac{\sum_{t\in\mathcal{F}_S}S_t}{|\mathcal{F}_S|}.
    \label{eq:structure-aggregation}
\end{equation}
A structure-valid frame contains at least one valid static or dynamic measurement; a measured score below $\tau_S$ still counts toward coverage. Multiplication by $C_S$ prevents a few assessable frames from representing an entire rollout. We report coverage separately because a low S can reflect structural failure or insufficient evidence.

The case-level OPIS score is a weighted arithmetic mean with nonnegative component weights $(w_P,w_I,w_S)$ satisfying $w_P+w_I+w_S=1$:
\begin{equation}
    \Score_{\mathrm{OPIS}}=100\left(w_P P+w_I I+w_S S\right).
    \label{eq:opis-score}
\end{equation}
The three components are evaluated on their respective evidence sets: confirmed missing instances are penalized through Presence, whereas Identity and Structure characterize the fidelity of instances with valid measurements. Consequently, the weighted OPIS score is a composite diagnostic rather than a joint probability that all input objects are preserved. Scores are macro-averaged over cases within each subcategory, then over subcategories within each domain, and finally over domains. This hierarchy prevents densely annotated cases or larger domains from dominating the benchmark. 

\section{Experiments}
\label{sec:experiments}

\subsection{Experimental Setup}
\label{sec:setup}
\textbf{Evaluator implementation.}
We use SAM~3.1~\citep{Carion2025SAM3} for instance segmentation, DINOv3~\citep{Simeoni2025DINOv3} for appearance embeddings, and MoGe-2~\citep{Wang2025MoGeTwo} with MASt3R~\citep{Leroy2024MASt3R} for input-grounded geometric evidence. The geometric reference uses up to 128 query points per instance. The association threshold is $0.35$, the ambiguity margin is $0.05$, and the normalized geometric tolerance is $\lambda_g=0.05$. Dynamic structure uses a hybrid pose--VLM pipeline: human and animal models based on ViTPose~\citep{Xu2022ViTPose} and ViTPose++~\citep{Xu2024ViTPosePlusPlus} measure landmark proportions, with VLM evidence used for other supported cases and pose fallback. The pose tolerance is $\tau=0.2$; VLM assessment uses at most 12 claims per object and a confidence threshold of $0.7$. Identity and Structure retain the mean valid object score only when all valid objects meet their respective thresholds. The identity and structure thresholds are $\tau_I=0.35$ and $\tau_S=0.35$, respectively, and P/I/S weights are fixed to $(w_P,w_I,w_S)=(0.2,0.4,0.4)$.

\textbf{Evaluation setup.}
We target approximately 10-second rollouts for each case, using the case-specific text instruction for image-to-video models and the corresponding structured camera task for camera-conditioned world models. Generation uses model-specific adapters and supported output resolutions and frame rates. Evaluation samples every eight frames. Temporal diagnostics use timestamps derived from the recorded output frame rate. Results are macro-averaged through the case--subcategory--domain hierarchy. We estimate 95\% confidence intervals using 2,000 bootstrap resamples of cases within each subcategory, retaining the same aggregation hierarchy.
We evaluate eight systems spanning two interfaces. The image-to-video group comprises Wan~3.0, Seedance~2.0~\citep{seedance2026seedance20advancingvideo}, Gemini~Omni~Flash~v1.1, and MiniMax~H3-Max-Turbo. The camera-conditioned group comprises SANA-WM~\citep{zhu2026sanawmefficientminutescaleworld}, LingBot-World~2.0~\citep{gao2026infiniteworldsversatileinteractions}, Matrix-Game~3.5~\citep{qian2026matrixgame35enhancingrealtime}, and Echo-WM-Flash~\citep{zhang2026echowmopenenterableomnimodal}.

\subsection{Main Results}
\label{sec:results}
Table~\ref{tab:main-results} reports the formal OPIS results. Seedance 2.0 and H3-Max-Turbo obtain the highest point estimates, at 56.01 and 55.57, respectively. Within the camera-conditioned group, Echo-WM-Flash has the highest point estimate at 51.42. We also report confidence intervals for the formal OPIS results.

\begin{table*}[t]
\centering
\tablecaption{Formal OPIS results. P, I, S, and structural frame coverage \(C_S\) are shown on a 0--100 scale. S already includes the coverage multiplier. All columns use the case--subcategory--domain hierarchy. OPIS reports a 95\% stratified case-bootstrap interval.}
\label{tab:main-results}
\begin{tabular}{lccccc}
\toprule
Model & P \(\uparrow\) & I \(\uparrow\) & S \(\uparrow\) & \(C_S\) (\%) & OPIS \(\uparrow\) {\scriptsize [95\% CI]} \\
\midrule
\multicolumn{6}{l}{\emph{Image-to-video}} \\
\midrule
Seedance 2.0 & 89.75 & 35.07 & \textbf{60.08} & \textbf{81.53} & \textbf{56.01} {\scriptsize [52.13, 59.69]} \\
H3-Max-Turbo & 93.04 & 36.77 & 55.64 & 76.77 & 55.57 {\scriptsize [52.20, 58.93]} \\
Wan-3.0 & 88.95 & 34.47 & 50.58 & 75.86 & 51.81 {\scriptsize [48.55, 55.03]} \\
Gemini-Omni-Flash & \textbf{93.49} & 30.03 & 49.30 & 70.94 & 50.43 {\scriptsize [47.13, 53.76]} \\
\midrule
\multicolumn{6}{l}{\emph{Camera-conditioned}} \\
\midrule
Echo-WM-Flash & 89.65 & 34.73 & 49.00 & 80.22 & 51.42 {\scriptsize [47.19, 55.62]} \\
LingBot-World 2.0 & 89.71 & 36.01 & 45.04 & 67.42 & 50.36 {\scriptsize [46.48, 54.25]} \\
Matrix-Game 3.5 & 85.63 & \textbf{36.88} & 45.85 & 77.01 & 50.22 {\scriptsize [45.73, 54.48]} \\
SANA-WM & 90.46 & 28.12 & 48.26 & 75.97 & 48.65 {\scriptsize [45.14, 52.29]} \\
\bottomrule
\end{tabular}
\vspace{\baselineskip}
\end{table*}

\textbf{Identity fidelity is substantially weaker than object accounting.}
Presence ranges from 85.63 to 93.49, while strict Identity ranges from 28.12 to 36.88. Across all rollouts, roughly half of the frames with appearance evidence contain at least one identity score below $\tau_I$. The gap is consistent across both model interfaces: systems can account for objects that remain observable, yet frequently fail to preserve the identity of the particular instances established in the input. Matrix-Game records the strongest Identity score (36.88), while Gemini-Omni-Flash records the strongest Presence score (93.49); neither dimension alone predicts the overall ranking. This separation is precisely what an object-centric memory benchmark should reveal and what a single scene-level score would hide.

\textbf{Presence is relatively mature, but not uniform across systems.}
All eight systems obtain high Presence scores (85.63--93.49), indicating that accounting for observable objects is comparatively mature, while still leaving a measurable gap between the strongest and weakest systems. Gemini-Omni-Flash has the highest Presence score (93.49), yet its Identity (30.03), Structure (49.30), and OPIS (50.43) scores are not among the strongest. The contrast shows that high object accounting does not imply faithful preservation of the particular input instances; overall memory quality depends on balancing Presence with Identity and Structure rather than optimizing P alone.

\textbf{Structural preservation requires both fidelity and evidence.}
Strong models not only preserve object structure more effectively, but also provide verifiable evidence of that preservation across a larger fraction of generated frames. Seedance 2.0 achieves the highest coverage-adjusted Structure score (60.08) and the broadest structural frame coverage (81.53\%). Across models, $C_S$ ranges from 67.42\% to 81.53\%, showing that structural memory depends both on preservation quality and evidence availability. Reporting Structure together with its coverage therefore distinguishes robust structural preservation from high scores supported by only a small number of assessable frames. Additional coverage diagnostics are provided in the appendix (Table~\ref{tab:coverage-detail}).

Per-subcategory results are included in Appendix~\ref{app:results}.

\subsection{Object-Centric Diagnostics}
\label{sec:diagnostics}

\begin{figure*}[t]
\centering
\includegraphics[width=\textwidth]{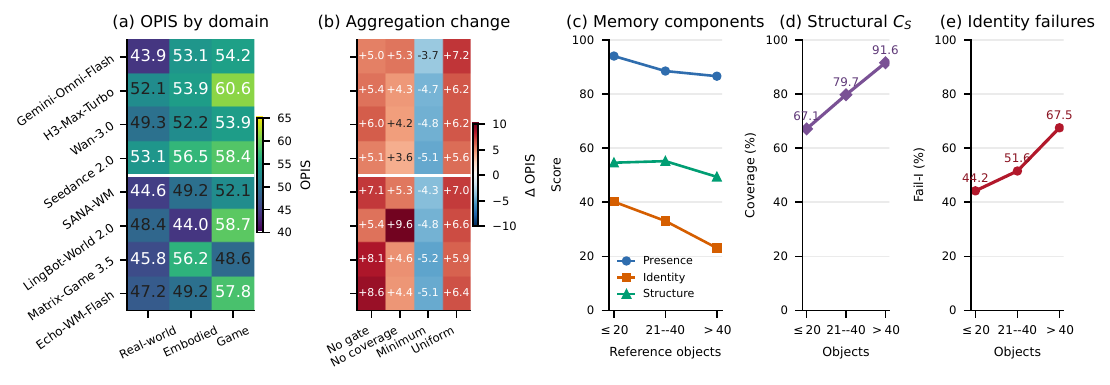}
\caption{Object-centric diagnostic views. Panel (a) reports OPIS across scene domains; each cell averages subcategories equally within a domain. Panel (b) shows the change in OPIS under aggregation variants relative to the strict score. Panels (c)--(e) show how reference-inventory size affects the memory components $P$, $I$, and $S$, structural evidence coverage $C_S$, and the fraction of identity-valid frames containing at least one strict-Identity threshold failure, respectively. Density values use the reported common-case diagnostic.}
\label{fig:domain-aggregation}
\label{fig:density-results}
\end{figure*}

\textbf{Multi-object load exposes a scaling challenge.} Figure~\ref{fig:density-results}(c) shows a clear scaling effect: scenes with at most 20 reference objects achieve an average strict Identity of 40.22, compared with 33.10 for 21--40 objects and 23.11 for more than 40 objects. The fraction of identity-valid frames containing at least one threshold failure rises from 44.24\% to 51.58\% and then 67.55\%, as shown in Figure~\ref{fig:density-results}(e). As the number of addressable instances grows, preserving every identity becomes substantially harder, revealing a central limitation of current world models in dense multi-object scenes.
\textbf{Cross-domain behavior.} Figure~\ref{fig:domain-aggregation}(a) shows a clear advantage on game worlds for seven of the eight systems, while Matrix-Game performs best on embodied scenes. The domain spread is substantial: LingBot-World reaches 58.74 on game worlds versus 44.00 on embodied scenes, whereas Matrix-Game reaches 56.21 on embodied scenes versus 48.65 on game worlds. These shifts show that memory performance depends strongly on scene composition, object configuration, and the type of visual evidence available, motivating evaluation across diverse domains rather than on a single scene family.

\subsection{Evaluator Validation and Ablations}
\label{sec:ablation}
\textbf{Aggregation.} Figure~\ref{fig:domain-aggregation}(b) recomputes scores from exactly the same archived object measurements. Removing the structural threshold raises scores by 5.05--8.60 points, because strong objects can then offset a weak one within the same frame. Removing the structural coverage multiplier raises scores by 3.62--9.61 points, with the largest increase for LingBot-World, whose structural coverage is lowest. Replacing the mean on passing frames with the minimum lowers scores by 3.71--5.24 points. Together, these ablations show that changing the aggregation rules produces consistent score shifts across models.

\textbf{Evaluator validity.} We assess structural-claim repeatability and agreement with manual review of association, visibility, identity, and structure. Table~\ref{tab:evaluator-validity} in Appendix~\ref{app:evaluator-validity} reports 90.5\% repeat agreement (181/200 claim pairs). Agreement with manual review is 92.7\% for association, 90.7\% for visibility, 86.0\% for identity, and 83.3\% for structure (150 decisions per dimension). Structure has the lowest agreement in this audit. Appendix~\ref{app:evaluator-validity} describes the evaluation units, agreement metric, and interpretation.

\section{Conclusion}
\label{sec:conclusion}
OPIS evaluates multi-object memory against a fixed initial observation through Presence, Identity, and Structure. Its strict frame-level criterion exposes failures that can be concealed by averaging over many objects, while separate coverage diagnostics identify limits of the available evidence. The current eight-model study reveals a substantial difference between object accounting and identity fidelity, as well as sensitivity to structural coverage. These findings motivate input-grounded object evaluation alongside broader measures of video quality and world-model capability.

\subsection*{AI Use Statement}

Generative AI tools were used to assist with manuscript drafting and editing,
dataset construction, and evaluation. Specifically, \texttt{gpt-image-2.5-sunburst}
generated 83 benchmark input images. Vision-language models assisted with
image selection, noun-phrase cleaning, task-specification drafting,
quality-control review, and selected dynamic-structure judgments. No
generative AI was used to replace human responsibility for scientific claims,
mathematical definitions, citations, final statistics, or release decisions;
other required AI-use categories are not applicable. All AI-assisted outputs were reviewed by the authors. We verified the
dataset inventory, masks, task specifications, evaluator implementation,
thresholds, aggregation rules, and reported statistics.

\subsection*{Ethics Statement}

This work does not involve human-subject experiments, participant
interaction, private data, or sensitive personal information. OPIS uses
licensed or attributed real-world, simulator, and game assets, together with
a limited set of AI-generated images. The benchmark is intended for academic evaluation of video world models and
does not provide instructions for harmful activity or surveillance. Its
results may reflect biases in source datasets, generated imagery,
segmentation models, and vision-language judgments.

\subsection*{Reproducibility Statement}

The paper specifies the input-grounded task, annotation pipeline, evaluator,
Presence--Identity--Structure metrics, aggregation hierarchy, thresholds,
sampling procedure, and bootstrap analysis. OPIS contains 500 cases and
12,672 annotated object instances across three domains and ten subcategories.
The appendices provide dataset composition, detailed results, ablations,
generation configurations, AI-image provenance, and evaluator-validity
checks. The evaluation code and configuration information are available at
\url{https://github.com/SSStarain/OPIS}; the dataset is available at
\url{https://huggingface.co/datasets/Kirito-Lab/OPIS-dataset}, subject to
applicable licenses. The reported
implementation uses SAM~3.1, DINOv3, MoGe-2, MASt3R, ViTPose/ViTPose++, and
the documented association, geometry, pose, identity, structure, and
aggregation settings, enabling independent reruns of the evaluator and
reported analyses.

\clearpage
\phantomsection
\addcontentsline{toc}{section}{References}
\bibliography{iclr2027_conference}

@article{Ha2018WorldModels,
  title={World models},
  author={Ha, David and Schmidhuber, J{\"u}rgen},
  journal={arXiv preprint arXiv:1803.10122},
  volume={2},
  number={3},
  pages={440},
  year={2018}
}

@article{Hafner2023DreamerV3,
  title={Mastering diverse domains through world models},
  author={Hafner, Danijar and Pasukonis, Jurgis and Ba, Jimmy and Lillicrap, Timothy},
  journal={arXiv preprint arXiv:2301.04104},
  year={2023}
}

@article{Hu2023GAIA1,
  title={Gaia-1: A generative world model for autonomous driving},
  author={Hu, Anthony and Russell, Lloyd and Yeo, Hudson and Murez, Zak and Fedoseev, George and Kendall, Alex and Shotton, Jamie and Corrado, Gianluca},
  journal={arXiv preprint arXiv:2309.17080},
  year={2023}
}

@inproceedings{Bruce2024Genie,
  title={Genie: Generative interactive environments},
  author={Bruce, Jake and Dennis, Michael D and Edwards, Ashley and Parker-Holder, Jack and Shi, Yuge and Hughes, Edward and Lai, Matthew and Mavalankar, Aditi and Steigerwald, Richie and Apps, Chris and others},
  booktitle={Forty-first international conference on machine learning},
  year={2024}
}

@article{Wu2025LongTermSpatialMemory,
  title={Video world models with long-term spatial memory},
  author={Wu, Tong and Yang, Shuai and Po, Ryan and Xu, Yinghao and Liu, Ziwei and Lin, Dahua and Wetzstein, Gordon},
  journal={Advances in Neural Information Processing Systems},
  volume={38},
  pages={49371--49393},
  year={2026}
}

@article{MIND2026,
  title={Mind: Benchmarking memory consistency and action control in world models},
  author={Ye, Yixuan and Lu, Xuanyu and Jiang, Yuxin and Gu, Yuchao and Zhao, Rui and Liang, Qiwei and Pan, Jiachun and Zhang, Fengda and Wu, Weijia and Wang, Alex Jinpeng},
  journal={arXiv preprint arXiv:2602.08025},
  year={2026}
}

@article{R2MBench2026,
  title={R2M-Bench: Evaluating Revisit Memory via Relative Consistency in Interactive Video World Models},
  author={Gu, Qiwen and Gao, Bingjie and Chen, Rui and Li, Geng and Li, Jifan and Wen, Qishuai and Niu, Li and Tang, Jing and Chu, Xiangxiang and Zhao, Junqiao},
  journal={arXiv preprint arXiv:2608.27328},
  year={2026}
}

@article{MBench2026,
  title={Mbench: A comprehensive benchmark on memory capability for video world models},
  author={Zhang, Shengjun and Zhang, Zhang and Huang, Simin and Tang, Zhenyu and Wang, Hanyang and Dai, Chensheng and Chen, Min and Li, Yifan and Li, Yuxin and Chen, Yingjie and others},
  journal={arXiv preprint arXiv:2606.00793},
  year={2026}
}

@article{PDI2026,
  title={Quantitative Video World Model Evaluation for Geometric-Consistency},
  author={Wu, Jiaxin and Pi, Yihao and Zhang, Yinling and Li, Yuheng and Zou, Xueyan},
  journal={arXiv preprint arXiv:2605.15185},
  year={2026}
}

@article{WorldTrace2026,
  title={Addressable Memory for Video World Models},
  author={Wu, Xindi and Elflein, Sven and Lucas, James and Russakovsky, Olga and Leal-Taix{\'e}, Laura and Paschalidou, Despoina and Lorraine, Jonathan and O{\v{s}}ep, Aljo{\v{s}}a},
  journal={arXiv preprint arXiv:2608.07408},
  year={2026}
}

@article{LatentSpatialMemory2026,
  title={Latent spatial memory for video world models},
  author={Wang, Weijie and Zhao, Haoyu and Yang, Yifan and Chen, Feng and Zhang, Zeyu and He, Yefei and Duan, Zicheng and Chen, Donny Y and Yang, Yuqing and Zhuang, Bohan},
  journal={arXiv preprint arXiv:2606.09828},
  year={2026}
}

@inproceedings{Huang2024VBench,
  author       = {Ziqi Huang and
                  Yinan He and
                  Jiashuo Yu and
                  Fan Zhang and
                  Chenyang Si and
                  Yuming Jiang and
                  Yuanhan Zhang and
                  Tianxing Wu and
                  Qingyang Jin and
                  Nattapol Chanpaisit and
                  Yaohui Wang and
                  Xinyuan Chen and
                  Limin Wang and
                  Dahua Lin and
                  Yu Qiao and
                  Ziwei Liu},
  title        = {VBench: Comprehensive Benchmark Suite for Video Generative Models},
  booktitle    = {{IEEE/CVF} Conference on Computer Vision and Pattern Recognition,
                  {CVPR} 2024, Seattle, WA, USA, June 16-22, 2024},
  pages        = {21807--21818},
  publisher    = {{IEEE}},
  year         = {2024},
  url          = {https://doi.org/10.1109/CVPR52733.2024.02060},
  doi          = {10.1109/CVPR52733.2024.02060},
  bibsource    = {dblp computer science bibliography, https://dblp.org}
}

@article{Zheng2025VBench20,
  author       = {Dian Zheng and
                  Ziqi Huang and
                  Hongbo Liu and
                  Kai Zou and
                  Yinan He and
                  Fan Zhang and
                  Yuanhan Zhang and
                  Jingwen He and
                  Wei{-}Shi Zheng and
                  Yu Qiao and
                  Ziwei Liu},
  title        = {VBench-2.0: Advancing Video Generation Benchmark Suite for Intrinsic
                  Faithfulness},
  journal      = {CoRR},
  volume       = {abs/2503.21755},
  year         = {2025},
  url          = {https://doi.org/10.48550/arXiv.2503.21755},
  doi          = {10.48550/ARXIV.2503.21755},
  eprinttype   = {arXiv},
  eprint       = {2503.21755},
  bibsource    = {dblp computer science bibliography, https://dblp.org}
}

@inproceedings{Wang2025MoGeTwo,
  author       = {Ruicheng Wang and
                  Sicheng Xu and
                  Yue Dong and
                  Yu Deng and
                  Jianfeng Xiang and
                  Zelong Lv and
                  Guangzhong Sun and
                  Xin Tong and
                  Jiaolong Yang},
  editor       = {Danielle Belgrave and
                  Cheng Zhang and
                  Laura N. Montoya and
                  Hsuan{-}Tien Lin and
                  Razvan Pascanu and
                  Piotr Koniusz and
                  Marzyeh Ghassemi and
                  Nancy Chen and
                  Iv{\'{a}}n Vladimir Meza Ru{\'{\i}}z and
                  Arturo Loaiza{-}Bonilla},
  title        = {MoGe-2: Accurate Monocular Geometry with Metric Scale and Sharp Details},
  booktitle    = {Advances in Neural Information Processing Systems 38: Annual Conference
                  on Neural Information Processing Systems 2025, NeurIPS 2025, San Diego,
                  CA, USA, December 2-7, 2025 / Mexico City, Mexico, November 30 - December
                  5, 2025},
  year         = {2025},
  url          = {http://papers.nips.cc/paper\_files/paper/2025/hash/336572db3e99930814d6b328d4220cb6-Abstract-Conference.html},
  bibsource    = {dblp computer science bibliography, https://dblp.org}
}

@inproceedings{Leroy2024MASt3R,
  author       = {Vincent Leroy and
                  Yohann Cabon and
                  J{\'{e}}r{\^{o}}me Revaud},
  editor       = {Ales Leonardis and
                  Elisa Ricci and
                  Stefan Roth and
                  Olga Russakovsky and
                  Torsten Sattler and
                  G{\"{u}}l Varol},
  title        = {Grounding Image Matching in 3D with MASt3R},
  booktitle    = {Computer Vision - {ECCV} 2024 - 18th European Conference, Milan, Italy,
                  September 29-October 4, 2024, Proceedings, Part {LXXII}},
  series       = {Lecture Notes in Computer Science},
  volume       = {15130},
  pages        = {71--91},
  publisher    = {Springer},
  year         = {2024},
  url          = {https://doi.org/10.1007/978-3-031-73220-1\_5},
  doi          = {10.1007/978-3-031-73220-1\_5},
  bibsource    = {dblp computer science bibliography, https://dblp.org}
}

@article{Simeoni2025DINOv3,
  author       = {Oriane Sim{\'{e}}oni and
                  Huy V. Vo and
                  Maximilian Seitzer and
                  Federico Baldassarre and
                  Maxime Oquab and
                  Cijo Jose and
                  Vasil Khalidov and
                  Marc Szafraniec and
                  Seung Eun Yi and
                  Micha{\"{e}}l Ramamonjisoa and
                  Francisco Massa and
                  Daniel Haziza and
                  Luca Wehrstedt and
                  Jianyuan Wang and
                  Timoth{\'{e}}e Darcet and
                  Th{\'{e}}o Moutakanni and
                  Leonel Sentana and
                  Claire Roberts and
                  Andrea Vedaldi and
                  Jamie Tolan and
                  John Brandt and
                  Camille Couprie and
                  Julien Mairal and
                  Herv{\'{e}} J{\'{e}}gou and
                  Patrick Labatut and
                  Piotr Bojanowski},
  title        = {DINOv3},
  journal      = {Trans. Mach. Learn. Res.},
  volume       = {2026},
  year         = {2026},
  url          = {https://openreview.net/forum?id=2NlGyqNjns},
  bibsource    = {dblp computer science bibliography, https://dblp.org}
}

@article{Carion2025SAM3,
  author       = {Nicolas Carion and
                  Laura Gustafson and
                  Yuan{-}Ting Hu and
                  Shoubhik Debnath and
                  Ronghang Hu and
                  Didac Suris and
                  Chaitanya Ryali and
                  Kalyan Vasudev Alwala and
                  Haitham Khedr and
                  Andrew Huang and
                  Jie Lei and
                  Tengyu Ma and
                  Baishan Guo and
                  Arpit Kalla and
                  Markus Marks and
                  Joseph Greer and
                  Meng Wang and
                  Peize Sun and
                  Roman R{\"{a}}dle and
                  Triantafyllos Afouras and
                  Effrosyni Mavroudi and
                  Katherine Xu and
                  Tsung{-}Han Wu and
                  Yu Zhou and
                  Liliane Momeni and
                  Rishi Hazra and
                  Shuangrui Ding and
                  Sagar Vaze and
                  Francois Porcher and
                  Feng Li and
                  Siyuan Li and
                  Aishwarya Kamath and
                  Ho Kei Cheng and
                  Piotr Doll{\'{a}}r and
                  Nikhila Ravi and
                  Kate Saenko and
                  Pengchuan Zhang and
                  Christoph Feichtenhofer},
  title        = {{SAM} 3: Segment Anything with Concepts},
  journal      = {CoRR},
  volume       = {abs/2511.16719},
  year         = {2025},
  url          = {https://doi.org/10.48550/arXiv.2511.16719},
  doi          = {10.48550/ARXIV.2511.16719},
  eprinttype   = {arXiv},
  eprint       = {2511.16719},
  bibsource    = {dblp computer science bibliography, https://dblp.org}
}

@inproceedings{Xu2022ViTPose,
  author       = {Yufei Xu and
                  Jing Zhang and
                  Qiming Zhang and
                  Dacheng Tao},
  editor       = {Sanmi Koyejo and
                  S. Mohamed and
                  A. Agarwal and
                  Danielle Belgrave and
                  K. Cho and
                  A. Oh},
  title        = {ViTPose: Simple Vision Transformer Baselines for Human Pose Estimation},
  booktitle    = {Advances in Neural Information Processing Systems 35: Annual Conference
                  on Neural Information Processing Systems 2022, NeurIPS 2022, New Orleans,
                  LA, USA, November 28 - December 9, 2022},
  year         = {2022},
  url          = {http://papers.nips.cc/paper\_files/paper/2022/hash/fbb10d319d44f8c3b4720873e4177c65-Abstract-Conference.html},
  bibsource    = {dblp computer science bibliography, https://dblp.org}
}

@article{Xu2024ViTPosePlusPlus,
  author       = {Yufei Xu and
                  Jing Zhang and
                  Qiming Zhang and
                  Dacheng Tao},
  title        = {ViTPose++: Vision Transformer for Generic Body Pose Estimation},
  journal      = {{IEEE} Trans. Pattern Anal. Mach. Intell.},
  volume       = {46},
  number       = {2},
  pages        = {1212--1230},
  year         = {2024},
  url          = {https://doi.org/10.1109/TPAMI.2023.3330016},
  doi          = {10.1109/TPAMI.2023.3330016},
  bibsource    = {dblp computer science bibliography, https://dblp.org}
}

@inproceedings{T2VCompBench2025,
  author       = {Kaiyue Sun and
                  Kaiyi Huang and
                  Xian Liu and
                  Yue Wu and
                  Zihan Xu and
                  Zhenguo Li and
                  Xihui Liu},
  title        = {T2V-CompBench: {A} Comprehensive Benchmark for Compositional Text-to-video
                  Generation},
  booktitle    = {{IEEE/CVF} Conference on Computer Vision and Pattern Recognition,
                  {CVPR} 2025, Nashville, TN, USA, June 11-15, 2025},
  pages        = {8406--8416},
  publisher    = {Computer Vision Foundation / {IEEE}},
  year         = {2025},
  url          = {https://openaccess.thecvf.com/content/CVPR2025/html/Sun\_T2V-CompBench\_A\_Comprehensive\_Benchmark\_for\_Compositional\_Text-to-video\_Generation\_CVPR\_2025\_paper.html},
  doi          = {10.1109/CVPR52734.2025.00787},
  bibsource    = {dblp computer science bibliography, https://dblp.org}
}

@article{Krishna2017VisualGenome,
  author       = {Ranjay Krishna and
                  Yuke Zhu and
                  Oliver Groth and
                  Justin Johnson and
                  Kenji Hata and
                  Joshua Kravitz and
                  Stephanie Chen and
                  Yannis Kalantidis and
                  Li{-}Jia Li and
                  David A. Shamma and
                  Michael S. Bernstein and
                  Li Fei{-}Fei},
  title        = {Visual Genome: Connecting Language and Vision Using Crowdsourced Dense
                  Image Annotations},
  journal      = {Int. J. Comput. Vis.},
  volume       = {123},
  number       = {1},
  pages        = {32--73},
  year         = {2017},
  url          = {https://doi.org/10.1007/s11263-016-0981-7},
  doi          = {10.1007/S11263-016-0981-7},
  bibsource    = {dblp computer science bibliography, https://dblp.org}
}

@inproceedings{Zhou2017ADE20K,
  author       = {Bolei Zhou and
                  Hang Zhao and
                  Xavier Puig and
                  Sanja Fidler and
                  Adela Barriuso and
                  Antonio Torralba},
  title        = {Scene Parsing through {ADE20K} Dataset},
  booktitle    = {2017 {IEEE} Conference on Computer Vision and Pattern Recognition,
                  {CVPR} 2017, Honolulu, HI, USA, July 21-26, 2017},
  pages        = {5122--5130},
  publisher    = {{IEEE} Computer Society},
  year         = {2017},
  url          = {https://doi.org/10.1109/CVPR.2017.544},
  doi          = {10.1109/CVPR.2017.544},
  bibsource    = {dblp computer science bibliography, https://dblp.org}
}

@inproceedings{Lin2014COCO,
  author       = {Tsung{-}Yi Lin and
                  Michael Maire and
                  Serge J. Belongie and
                  James Hays and
                  Pietro Perona and
                  Deva Ramanan and
                  Piotr Doll{\'{a}}r and
                  C. Lawrence Zitnick},
  editor       = {David J. Fleet and
                  Tom{\'{a}}s Pajdla and
                  Bernt Schiele and
                  Tinne Tuytelaars},
  title        = {Microsoft {COCO:} Common Objects in Context},
  booktitle    = {Computer Vision - {ECCV} 2014 - 13th European Conference, Zurich,
                  Switzerland, September 6-12, 2014, Proceedings, Part {V}},
  series       = {Lecture Notes in Computer Science},
  volume       = {8693},
  pages        = {740--755},
  publisher    = {Springer},
  year         = {2014},
  url          = {https://doi.org/10.1007/978-3-319-10602-1\_48},
  doi          = {10.1007/978-3-319-10602-1\_48},
  bibsource    = {dblp computer science bibliography, https://dblp.org}
}

@inproceedings{Nagaraja2016ReferringExpressions,
  author       = {Licheng Yu and
                  Patrick Poirson and
                  Shan Yang and
                  Alexander C. Berg and
                  Tamara L. Berg},
  editor       = {Bastian Leibe and
                  Jiri Matas and
                  Nicu Sebe and
                  Max Welling},
  title        = {Modeling Context in Referring Expressions},
  booktitle    = {Computer Vision - {ECCV} 2016 - 14th European Conference, Amsterdam,
                  The Netherlands, October 11-14, 2016, Proceedings, Part {II}},
  series       = {Lecture Notes in Computer Science},
  volume       = {9906},
  pages        = {69--85},
  publisher    = {Springer},
  year         = {2016},
  url          = {https://doi.org/10.1007/978-3-319-46475-6\_5},
  doi          = {10.1007/978-3-319-46475-6\_5},
  bibsource    = {dblp computer science bibliography, https://dblp.org}
}

@inproceedings{Walke2023BridgeData,
  author       = {Homer Rich Walke and
                  Kevin Black and
                  Tony Z. Zhao and
                  Quan Vuong and
                  Chongyi Zheng and
                  Philippe Hansen{-}Estruch and
                  Andre Wang He and
                  Vivek Myers and
                  Moo Jin Kim and
                  Max Du and
                  Abraham Lee and
                  Kuan Fang and
                  Chelsea Finn and
                  Sergey Levine},
  editor       = {Jie Tan and
                  Marc Toussaint and
                  Kourosh Darvish},
  title        = {BridgeData {V2:} {A} Dataset for Robot Learning at Scale},
  booktitle    = {Conference on Robot Learning, CoRL 2023, 6-9 November 2023, Atlanta,
                  GA, {USA}},
  series       = {Proceedings of Machine Learning Research},
  volume       = {229},
  pages        = {1723--1736},
  publisher    = {{PMLR}},
  year         = {2023},
  url          = {https://proceedings.mlr.press/v229/walke23a.html},
  bibsource    = {dblp computer science bibliography, https://dblp.org}
}

@inproceedings{Fan2022MineDojo,
  author       = {Linxi Fan and
                  Guanzhi Wang and
                  Yunfan Jiang and
                  Ajay Mandlekar and
                  Yuncong Yang and
                  Haoyi Zhu and
                  Andrew Tang and
                  De{-}An Huang and
                  Yuke Zhu and
                  Anima Anandkumar},
  editor       = {Sanmi Koyejo and
                  S. Mohamed and
                  A. Agarwal and
                  Danielle Belgrave and
                  K. Cho and
                  A. Oh},
  title        = {MineDojo: Building Open-Ended Embodied Agents with Internet-Scale
                  Knowledge},
  booktitle    = {Advances in Neural Information Processing Systems 35: Annual Conference
                  on Neural Information Processing Systems 2022, NeurIPS 2022, New Orleans,
                  LA, USA, November 28 - December 9, 2022},
  year         = {2022},
  url          = {http://papers.nips.cc/paper\_files/paper/2022/hash/74a67268c5cc5910f64938cac4526a90-Abstract-Datasets\_and\_Benchmarks.html},
  bibsource    = {dblp computer science bibliography, https://dblp.org}
}

@article{Kolve2017AI2THOR,
  author       = {Eric Kolve and
                  Roozbeh Mottaghi and
                  Daniel Gordon and
                  Yuke Zhu and
                  Abhinav Gupta and
                  Ali Farhadi},
  title        = {{AI2-THOR:} An Interactive 3D Environment for Visual {AI}},
  journal      = {CoRR},
  volume       = {abs/1712.05474},
  year         = {2017},
  url          = {http://arxiv.org/abs/1712.05474},
  eprinttype   = {arXiv},
  eprint       = {1712.05474},
  bibsource    = {dblp computer science bibliography, https://dblp.org}
}

@incollection{Kuhn1955Hungarian,
  author       = {Harold W. Kuhn},
  editor       = {Michael J{\"{u}}nger and
                  Thomas M. Liebling and
                  Denis Naddef and
                  George L. Nemhauser and
                  William R. Pulleyblank and
                  Gerhard Reinelt and
                  Giovanni Rinaldi and
                  Laurence A. Wolsey},
  title        = {The Hungarian Method for the Assignment Problem},
  booktitle    = {50 Years of Integer Programming 1958-2008 - From the Early Years to
                  the State-of-the-Art},
  pages        = {29--47},
  publisher    = {Springer},
  year         = {2010},
  url          = {https://doi.org/10.1007/978-3-540-68279-0\_2},
  doi          = {10.1007/978-3-540-68279-0\_2},
  bibsource    = {dblp computer science bibliography, https://dblp.org}
}

@misc{ying2026wbenchcomprehensivemultiturnbenchmark,
      title={WBench: A Comprehensive Multi-turn Benchmark for Interactive Video World Model Evaluation}, 
      author={Kaining Ying and Hengrui Hu and Siyu Ren and Jiamu Li and Fengjiao Chen and Ziwen Wang and Xuezhi Cao and Xunliang Cai and Henghui Ding},
      year={2026},
      eprint={2605.25874},
      archivePrefix={arXiv},
      primaryClass={cs.CV},
      url={https://arxiv.org/abs/2605.25874}, 
}

@misc{duan2025worldscoreunifiedevaluationbenchmark,
      title={WorldScore: A Unified Evaluation Benchmark for World Generation}, 
      author={Haoyi Duan and Hong-Xing Yu and Sirui Chen and Li Fei-Fei and Jiajun Wu},
      year={2025},
      eprint={2504.00983},
      archivePrefix={arXiv},
      primaryClass={cs.GR},
      url={https://arxiv.org/abs/2504.00983}, 
}

@misc{Carion2020DETR,
      title={End-to-End Object Detection with Transformers}, 
      author={Nicolas Carion and Francisco Massa and Gabriel Synnaeve and Nicolas Usunier and Alexander Kirillov and Sergey Zagoruyko},
      year={2020},
      eprint={2005.12872},
      archivePrefix={arXiv},
      primaryClass={cs.CV},
      url={https://arxiv.org/abs/2005.12872}, 
}

@misc{seedance2026seedance20advancingvideo,
      title={Seedance 2.0: Advancing Video Generation for World Complexity}, 
      author={Team Seedance and De Chen and Liyang Chen and Xin Chen and Ying Chen and Zhuo Chen and Zhuowei Chen and Feng Cheng and Tianheng Cheng and Yufeng Cheng and Mojie Chi and Xuyan Chi and Jian Cong and Qinpeng Cui and Fei Ding and Qide Dong and Yujiao Du and Haojie Duanmu and Junliang Fan and Jiarui Fang and Jing Fang and Zetao Fang and Chengjian Feng and Yu Gao and Diandian Gu and Dong Guo and Hanzhong Guo and Qiushan Guo and Boyang Hao and Hongxiang Hao and Haoxun He and Jiaao He and Qian He and Tuyen Hoang and Heng Hu and Ruoqing Hu and Yuxiang Hu and Jiancheng Huang and Weilin Huang and Zhaoyang Huang and Zhongyi Huang and Jishuo Jin and Ming Jing and Ashley Kim and Shanshan Lao and Yichong Leng and Bingchuan Li and Gen Li and Haifeng Li and Huixia Li and Jiashi Li and Ming Li and Xiaojie Li and Xingxing Li and Yameng Li and Yiying Li and Yu Li and Yueyan Li and Chao Liang and Han Liang and Jianzhong Liang and Ying Liang and Wang Liao and J. H. Lien and Shanchuan Lin and Xi Lin and Feng Ling and Yue Ling and Fangfang Liu and Jiawei Liu and Jihao Liu and Jingtuo Liu and Shu Liu and Sichao Liu and Wei Liu and Xue Liu and Zuxi Liu and Ruijie Lu and Lecheng Lyu and Jingting Ma and Tianxiang Ma and Xiaonan Nie and Jingzhe Ning and Junjie Pan and Xitong Pan and Ronggui Peng and Xueqiong Qu and Yuxi Ren and Yuchen Shen and Guang Shi and Lei Shi and Yinglong Song and Fan Sun and Li Sun and Renfei Sun and Wenjing Tang and Boyang Tao and Zirui Tao and Dongliang Wang and Feng Wang and Hulin Wang and Ke Wang and Qingyi Wang and Rui Wang and Shuai Wang and Shulei Wang and Weichen Wang and Xuanda Wang and Yanhui Wang and Yue Wang and Yuping Wang and Yuxuan Wang and Zijie Wang and Ziyu Wang and Guoqiang Wei and Meng Wei and Di Wu and Guohong Wu and Hanjie Wu and Huachao Wu and Jian Wu and Jie Wu and Ruolan Wu and Shaojin Wu and Xiaohu Wu and Xinglong Wu and Yonghui Wu and Ruiqi Xia and Xin Xia and Xuefeng Xiao and Shuang Xu and Bangbang Yang and Jiaqi Yang and Runkai Yang and Tao Yang and Yihang Yang and Zhixian Yang and Ziyan Yang and Fulong Ye and Bingqian Yi and Xing Yin and Yongbin You and Linxiao Yuan and Weihong Zeng and Xuejiao Zeng and Yan Zeng and Siyu Zhai and Zhonghua Zhai and Bowen Zhang and Chenlin Zhang and Heng Zhang and Jun Zhang and Manlin Zhang and Peiyuan Zhang and Shuo Zhang and Xiaohe Zhang and Xiaoying Zhang and Xinyan Zhang and Xinyi Zhang and Yichi Zhang and Zixiang Zhang and Haiyu Zhao and Huating Zhao and Liming Zhao and Yian Zhao and Guangcong Zheng and Jianbin Zheng and Xiaozheng Zheng and Zerong Zheng and Kuan Zhu and Feilong Zuo},
      year={2026},
      eprint={2604.14148},
      archivePrefix={arXiv},
      primaryClass={cs.CV},
      url={https://arxiv.org/abs/2604.14148}, 
}

@misc{gao2026infiniteworldsversatileinteractions,
      title={Infinite Worlds with Versatile Interactions}, 
      author={Zelin Gao and Qiuyu Wang and Jiapeng Zhu and Jingye Chen and Zichen Liu and Qingyan Bai and Jiahao Wang and Yufeng Yuan and Hanlin Wang and Yichong Lu and Ka Leong Cheng and Haojie Zhang and Jian Gao and Tianrui Feng and Yuzheng Liu and Yao Yao and Yinghao Xu and Xing Zhu and Yujun Shen and Hao Ouyang},
      year={2026},
      eprint={2607.07534},
      archivePrefix={arXiv},
      primaryClass={cs.CV},
      url={https://arxiv.org/abs/2607.07534}, 
}

@misc{qian2026matrixgame35enhancingrealtime,
      title={Matrix-Game 3.5: Enhancing Real-Time Streaming Interactive World Models with Patch Memory}, 
      author={Runjia Qian and Zile Wang and Jihai Zhang and Kai Zou and Wei Yu and Jiaxing Li and Zexiang Liu and Yaokun Li and Fei Kang and Kaichen Huang and Mengyin An and Haobo Zhang and Biao Jiang and Jiahua Wang and Haofeng Sun and Yang Liu and Yangguang Li},
      year={2026},
      eprint={2608.29910},
      archivePrefix={arXiv},
      primaryClass={cs.CV},
      url={https://arxiv.org/abs/2608.29910}, 
}

@misc{zhang2026echowmopenenterableomnimodal,
      title={EchoWM: Open and Enterable Omnimodal World Models}, 
      author={Songchun Zhang and Yaowei Li and Junhao Zhuang and Weiyang Jin and Haoyu Wang and Xin Lu and Yilang Sun and Shiyi Zhang and Haoran Li and Xiaoxiao Ma and Yuming Li and Yijun Liu and Yaofeng Su and Yanwen Ma and Haoyu Wu and Zihan Su and Yue Ma and Lvmin Zhang and Haoyang Huang and Zeyue Xue and Anyi Rao and Nan Duan},
      year={2026},
      eprint={2608.23189},
      archivePrefix={arXiv},
      primaryClass={cs.CV},
      url={https://arxiv.org/abs/2608.23189}, 
}

@misc{zhu2026sanawmefficientminutescaleworld,
      title={SANA-WM: Efficient Minute-Scale World Modeling with Hybrid Linear Diffusion Transformer}, 
      author={Haoyi Zhu and Haozhe Liu and Yuyang Zhao and Tian Ye and Junsong Chen and Jincheng Yu and Tong He and Song Han and Enze Xie},
      year={2026},
      eprint={2605.15178},
      archivePrefix={arXiv},
      primaryClass={cs.CV},
      url={https://arxiv.org/abs/2605.15178}, 
}

@misc{sapoutzoglou2026industryshapesrgbdbenchmarkdataset,
      title={IndustryShapes: An RGB-D Benchmark dataset for 6D object pose estimation of industrial assembly components and tools}, 
      author={Panagiotis Sapoutzoglou and Orestis Vaggelis and Athina Zacharia and Evangelos Sartinas and Maria Pateraki},
      year={2026},
      eprint={2602.05555},
      archivePrefix={arXiv},
      primaryClass={cs.CV},
      url={https://arxiv.org/abs/2602.05555}, 
}
\bibliographystyle{unsrtnat}

\clearpage
\appendix

\section{Code and Data Availability}
\label{app:code-data-availability}
The OPIS evaluation code is available at \url{https://github.com/SSStarain/OPIS},
and the dataset is available at
\url{https://huggingface.co/datasets/Kirito-Lab/OPIS-dataset}.

\section{Limitations}
\label{sec:limitations}
OPIS measures observable preservation of input objects, rather than internal memory mechanisms, physical causality, or complete world-model competence. Its monocular reference provides estimated visible geometry, not complete 3D ground truth. Segmentation, association, camera estimation, and pose or VLM judgments can all limit the evidence. Strict frame scoring amplifies individual failures, including evaluator errors, and is more demanding in frames with more evaluable objects. The maximum video-generation length supported by some models also limits the study, so long-rollout behavior is not evaluated.

\section{Dataset Composition}
\label{app:dataset-composition}
OPIS contains 12,672 object instances across 500 cases, with 6,128 assigned to static-geometry evaluation and 6,544 to dynamic-structure evaluation (Table~\ref{tab:dataset-domains}).

\begin{table}[t]
\centering
\tablecaption{Composition of OPIS. Static and dynamic columns indicate the evaluation track assigned to each reference instance.}
\label{tab:dataset-domains}
\begin{tabular}{ccccc}
\toprule
Domain & Scenes & Instances & Static & Dynamic \\
\midrule
Real-world scenes & 200 & 8,082 & 3,565 & 4,517 \\
Embodied / robotic & 150 & 1,932 & 864 & 1,068 \\
Game worlds & 150 & 2,658 & 1,699 & 959 \\
\midrule
Total & 500 & 12,672 & 6,128 & 6,544 \\
\bottomrule
\end{tabular}
\vspace{\baselineskip}
\end{table}

\section{Detailed Results and Reproducibility}
\label{app:results}
This section provides the numeric tables corresponding to the main-text figures (Table~\ref{tab:density-detail}, Table~\ref{tab:aggregation-ablation-detail}, and Table~\ref{tab:domain-results-detail}), and additional per-subcategory scores (Table~\ref{tab:subcat1}, Table~\ref{tab:subcat2}, and Table~\ref{tab:subcat3}). They are retained for reproducibility and detailed lookup.

The reported confidence intervals are computed only for the main results in Table~\ref{tab:main-results}. For each model, we perform a stratified case-level bootstrap: within each of its ten subcategories, we resample the available cases with replacement, recompute the subcategory--domain--overall hierarchy using the same aggregation rules, and take the 2.5th and 97.5th percentiles over 2,000 replicates. This preserves the intended domain weights while propagating case-level variation through the reported hierarchy.

\begin{table}[t]
\centering
\tablecaption{Multi-object load across reference-inventory sizes. All score and rate columns use a 0--100 scale.}
\label{tab:density-detail}
\begin{tabular}{cccccc}
\toprule
Reference objects & P & I & S & Fail-I (\%) & $C_S$ (\%) \\
\midrule
$\leq 20$ & 94.06 & 40.22 & 54.65 & 44.24 & 67.14 \\
21--40 & 88.52 & 33.10 & 55.22 & 51.58 & 79.74 \\
$>40$ & 86.63 & 23.11 & 49.42 & 67.55 & 91.60 \\
\bottomrule
\end{tabular}
\vspace{\baselineskip}
\end{table}

\begin{table}[t]
\centering
\tablecaption{Aggregation sensitivity on the reported evaluation set. No gate removes only the $\tau_S$ structural threshold; no coverage removes only $C_S$; minimum replaces the within-frame mean for both I and S after thresholding; uniform uses equal P/I/S weights. All perception and association evidence is held fixed.}
\label{tab:aggregation-ablation-detail}
\begin{tabular}{cccccc}
\toprule
Model & Strict & No gate & No coverage & Minimum & Uniform \\
\midrule
Gemini-Omni-Flash & 50.43 & 55.48 & 55.75 & 46.72 & 57.61 \\
H3-Max-Turbo & 55.57 & 60.99 & 59.87 & 50.90 & 61.81 \\
Wan-3.0 & 51.81 & 57.84 & 55.97 & 47.00 & 58.00 \\
Seedance 2.0& 56.01 & 61.08 & 59.62 & 50.93 & 61.63 \\
SANA-WM & 48.65 & 55.70 & 53.96 & 44.38 & 55.62 \\
LingBot-World 2.0 & 50.36 & 55.78 & 59.97 & 45.60 & 56.92 \\
Matrix-Game 3.5 & 50.22 & 58.27 & 54.78 & 44.97 & 56.12 \\
Echo-WM-Flash & 51.42 & 60.02 & 55.86 & 46.36 & 57.80 \\
\bottomrule
\end{tabular}
\vspace{\baselineskip}
\end{table}

\begin{table}[t]
\centering
\tablecaption{OPIS by scene domain. Each domain score averages its subcategories equally.}
\label{tab:domain-results-detail}
\begin{tabular}{cccc}
\toprule
Model & Real-world & Embodied & Game \\
\midrule
Gemini-Omni-Flash & 43.92 & 53.14 & 54.23 \\
H3-Max-Turbo & 52.13 & 53.92 & 60.65 \\
Wan-3.0 & 49.31 & 52.19 & 53.93 \\
Seedance 2.0& 53.13 & 56.54 & 58.35 \\
SANA-WM & 44.62 & 49.21 & 52.11 \\
LingBot-World 2.0 & 48.35 & 44.00 & 58.74 \\
Matrix-Game 3.5 & 45.79 & 56.21 & 48.65 \\
Echo-WM-Flash & 47.22 & 49.23 & 57.82 \\
\bottomrule
\end{tabular}
\vspace{\baselineskip}
\end{table}

\begin{table}[t]
\centering
\tablecaption{Real-world subcategories: OPIS score.}
\label{tab:subcat1}
\begin{tabular}{ccccc}
\toprule
Model & Home & Natural & Public indoor & Urban \\
\midrule
Gemini-Omni-Flash & 36.89 & 54.23 & 32.57 & 51.99 \\
H3-Max-Turbo & 41.02 & 70.88 & 45.81 & 50.83 \\
Wan-3.0 & 29.17 & 70.69 & 43.79 & 53.61 \\
Seedance 2.0& 35.33 & 69.12 & 47.73 & 60.34 \\
SANA-WM & 23.11 & 54.68 & 46.86 & 53.83 \\
LingBot-World 2.0 & 27.55 & 66.33 & 40.68 & 58.84 \\
Matrix-Game 3.5 & 26.12 & 62.54 & 40.70 & 53.78 \\
Echo-WM-Flash & 32.24 & 64.36 & 41.68 & 50.62 \\
\bottomrule
\end{tabular}
\vspace{\baselineskip}
\end{table}
\begin{table}[t]
\centering
\tablecaption{Embodied subcategories: OPIS score.}
\label{tab:subcat2}
\begin{tabular}{cccc}
\toprule
Model & Industrial & Laboratory & Simulated \\
\midrule
Gemini-Omni-Flash & 43.50 & 65.41 & 50.50 \\
H3-Max-Turbo & 45.17 & 65.63 & 50.95 \\
Wan-3.0 & 49.05 & 64.79 & 42.71 \\
Seedance 2.0& 51.73 & 67.21 & 50.68 \\
SANA-WM & 44.83 & 62.44 & 40.35 \\
LingBot-World 2.0 & 40.70 & 48.72 & 42.57 \\
Matrix-Game 3.5 & 51.35 & 68.38 & 48.89 \\
Echo-WM-Flash & 45.70 & 68.08 & 33.90 \\
\bottomrule
\end{tabular}
\vspace{\baselineskip}
\end{table}
\begin{table}[t]
\centering
\tablecaption{Game subcategories: OPIS score.}
\label{tab:subcat3}
\begin{tabular}{cccc}
\toprule
Model & Cartoon & Pixel & Realistic \\
\midrule
Gemini-Omni-Flash & 54.68 & 55.03 & 52.99 \\
H3-Max-Turbo & 63.11 & 58.09 & 60.76 \\
Wan-3.0 & 45.49 & 56.30 & 60.00 \\
Seedance 2.0& 54.24 & 52.10 & 68.71 \\
SANA-WM & 46.97 & 54.72 & 54.64 \\
LingBot-World 2.0 & 56.06 & 66.83 & 53.32 \\
Matrix-Game 3.5 & 46.67 & 41.01 & 58.28 \\
Echo-WM-Flash & 52.57 & 61.78 & 59.12 \\
\bottomrule
\end{tabular}
\vspace{\baselineskip}
\end{table}

\begin{table}[t]
\centering
\tablecaption{Coverage audit. Percentage columns follow hierarchical averaging. No-S is a rollout count.}
\label{tab:coverage-detail}
\begin{tabular}{ccccccc}
\toprule
Model & $C_P$ & $C_I$ & Static & Dynamic & Unknown & No-S \\
\midrule
Gemini-Omni-Flash & 97.39 & 97.32 & 17.35 & 67.33 & 24.27 & 3 \\
H3-Max-Turbo & 97.46 & 97.46 & 19.68 & 74.23 & 26.50 & 3 \\
Wan-3.0 & 98.68 & 98.52 & 23.64 & 72.41 & 25.19 & 3 \\
Seedance 2.0& 97.20 & 97.20 & 18.99 & 79.19 & 28.59 & 2 \\
SANA-WM & 96.88 & 96.88 & 23.73 & 75.02 & 23.79 & 3 \\
LingBot-World 2.0 & 91.88 & 91.88 & 24.59 & 65.23 & 35.62 & 1 \\
Matrix-Game 3.5 & 90.37 & 90.22 & 28.77 & 73.40 & 33.21 & 2 \\
Echo-WM-Flash & 99.26 & 99.26 & 27.61 & 76.83 & 18.78 & 2 \\
\bottomrule
\end{tabular}
\vspace{\baselineskip}
\end{table}

\section{Temporal Memory Diagnostics}
\label{app:temporal}
\begin{figure}[t]
\centering
\includegraphics[width=\linewidth]{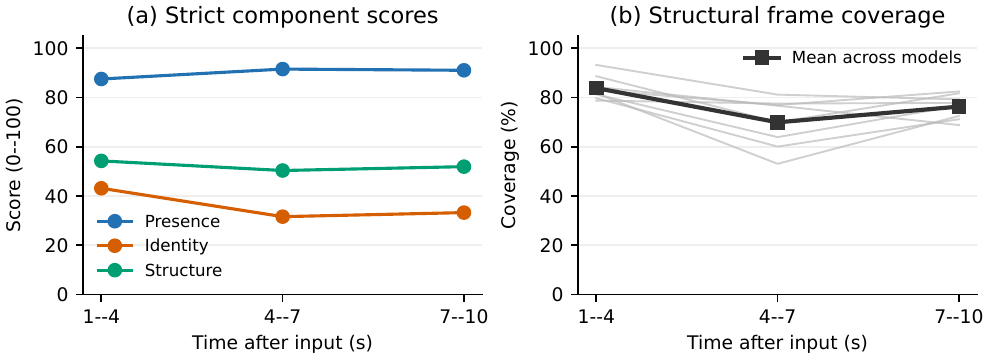}
\caption{Temporal memory diagnostics. Each interval is scored independently using the same strict rules. Left: component scores, averaged equally across models after hierarchical aggregation. Right: structural frame coverage; gray curves show individual models and black shows their mean.}
\label{fig:temporal-diagnostics}
\end{figure}

The temporal diagnostic averages each model's hierarchy-weighted scores equally across the matched evaluation set (Figure~\ref{fig:temporal-diagnostics}). Strict Identity drops from 43.13 in the 1--4\,s interval to 31.60 in 4--7\,s, before reaching 33.24 in 7--10\,s. Structure is more stable, moving from 54.23 to 50.33 and 51.90, while structural frame coverage changes from 83.79\% to 69.89\% and then 76.27\%. The result identifies identity preservation as the most fragile component of multi-object memory as generation proceeds.

\section{Additional Protocols}
\label{app:additional}
\subsection{Generation and evaluator configurations.}
Table~\ref{tab:configurations} records the output format and sampling density in the evaluation. Duration is measured from the output file, which can differ slightly from the requested 10 seconds. All models use an eight-frame stride after the first second, so their effective sampling rates differ. Temporal plots use seconds rather than sample indices. Evaluator thresholds are given in Section~\ref{sec:setup}; per-run manifests retain generation settings and judge configurations.

\subsection{AI-Generated Dataset Images and VLM Quality Control}
\label{app:ai-generated}

To supplement the collected real-world, embodied, and game scenes, we include a
small procedurally specified subset of AI-generated initial images.  The final
subset contains 83 images generated with \texttt{gpt-image-2.5-sunburst}: 21
\textit{home indoor}, 30 \textit{natural outdoor}, 17 \textit{public indoor},
and 15 \textit{urban outdoor} cases.  These images are used as benchmark
initial observations.

\begin{table}[t]
\centering
\tablecaption{AI-generated images in the real-world portion of OPIS.}
\label{tab:ai-generated-counts}
\begin{tabular}{cc}
\toprule
Scene subcategory & Number of cases \\
\midrule
Home indoor & 21 \\
Natural outdoor & 30 \\
Public indoor & 17 \\
Urban outdoor & 15 \\
\midrule
Total & 83 \\
\bottomrule
\end{tabular}
\vspace{\baselineskip}
\end{table}

\textbf{Image generation.} For each subcategory, the generator instantiates a setting description and a
composition variant (for example, a kitchen, a botanical garden path, a
library, or a transit plaza).  A variation index is included to request a
distinct location and layout across cases.  The prompt is designed for
image-to-video conditioning: it requests a single coherent 16:9 view with a
layered foreground and middle ground, at least 12 clearly visible object
instances, varied depth and overlap, and surfaces and edges that can support
subsequent motion.  It also excludes readable text, logos, watermarks, UI
elements, borders, duplicated or malformed objects, heavy blur, and scenes
dominated by undifferentiated background scenery.  The reusable template is
shown below.

\begin{tcolorbox}[
    title=\textbf{Image-generation prompt template},
    fonttitle=\bfseries,
    breakable,
    fontupper=\small,
    before upper={\setlength{\parindent}{0pt}\setlength{\parskip}{0.6\baselineskip}}
]
Create \texttt{\textless{}setting\textgreater{}}.

Scene category: \texttt{\textless{}subcategory\textgreater{}}.

Scene variation index: \texttt{\textless{}variation-index\textgreater{}}.  Use this as a distinct location
and layout from every other generated case; do not reuse a generic view.

Composition variation: \texttt{\textless{}composition-variant\textgreater{}}.

This is a benchmark initial frame for image-to-video generation.  Compose a
single coherent 16:9 camera view with a rich, layered foreground and middle
ground.  Include at least 12 clearly visible, distinct foreground object
instances with varied sizes and depths, including several overlapping but
individually recognizable objects.  Make the scene useful for motion: objects
should have clear surfaces, edges, depth ordering, and plausible spatial
relationships.

Use natural photographic lighting, sharp focus on the main scene, realistic
materials, balanced composition, and high detail.  Keep important objects away
from the extreme edges.  Do not make a collage, catalog, isolated studio
arrangement, empty landscape, or minimalist scene.

No text.  Exclude readable text, logos, watermarks, UI, borders, subtitles,
artificial labels, duplicated objects, malformed objects, extra limbs, and
heavy blur.  Do not let sky, grass, pavement, walls, or other scenery
dominate the frame; they may appear only as supporting context behind many
concrete foreground objects.  The result must look like one real photograph.
\end{tcolorbox}

Image requests use one sample per prompt (\texttt{n=1}), the high-quality
setting, and a 1536$\times$1024 JPEG response.  Each response is converted to
RGB and normalized to a 1280$\times$720 JPEG before it enters the benchmark.
The generation model, prompt revision, variation index, output dimensions, and
response metadata are retained with each case for provenance.

\textbf{VLM review and acceptance.} Every generated candidate is reviewed with \texttt{gemini-2.5-flash} using
the image and a structured instruction.  The reviewer returns one JSON object
containing (i) the final list of short, lowercase noun phrases for concrete
foreground objects, (ii) phrases removed from or added to the candidate list,
(iii) a list with one entry per distinct visible foreground instance, (iv) a
quality score from 1 to 5, and (v) quality flags.  The review removes
background-only surfaces and scenery, as well as text and HUD/UI elements;
multiple instances of the same category remain separate in the instance list.
The resulting noun-phrase inventory is used for the generated-image metadata,
while dense object-level masks and other benchmark annotations are produced by
the general annotation pipeline.

\begin{tcolorbox}[
    title=\textbf{VLM review prompt template},
    fonttitle=\bfseries,
    breakable,
    fontupper=\small,
    before upper={\setlength{\parindent}{0pt}\setlength{\parskip}{0.6\baselineskip}}
]
Inspect the image and clean the candidate noun-phrase list.  Keep only concrete
visible foreground objects, characters, and useful object parts.  Add visible
foreground objects that are missing from the list, but do not invent objects.

Remove background-only scenery and surfaces, readable text, logos, watermarks,
HUDs, menus, buttons, icons, counters, and other interface elements.

Return one JSON object with the fields
\texttt{noun\_phrases}, \texttt{removed\_noun\_phrases},
\texttt{added\_noun\_phrases}, \texttt{foreground\_object\_instances},
\texttt{quality\_score}, and \texttt{quality\_flags}.

Use short lowercase English noun phrases without leading articles.  List one
entry per distinct visible foreground instance in \texttt{foreground\_object\_instances}.

Assign
\texttt{quality\_score} from 1 (unusable) to 5 (clear), and return an empty
\texttt{quality\_flags} list when no quality issue is present.
\end{tcolorbox}

A candidate is retained only if the review reports at least 10 foreground
object instances, a quality score of at least 4/5, at least 5 usable noun
phrases, and no quality flags.  This gate is intended to enforce scene
complexity and visual usability for an image-to-video initial frame; it does
not replace the object-level evaluation protocol.  Per-case provenance records
the generation and review models, prompt versions, extracted object phrases,
instance count, quality score, and any review decisions.  AI-generated images
are not assumed to be redistributable solely because they were generated; any
release must separately verify the applicable usage and licensing conditions.

\begin{table}[t]
\centering
\tablecaption{Recorded generation formats and evaluation sampling. The Samples column counts evaluated frames per rollout.}
\label{tab:configurations}
\begin{tabular}{cccccc}
\toprule
Model & Resolution & FPS & Duration (s) & Stride & Samples \\
\midrule
Wan-3.0 & $854\times 480$ & 30 & 10.000 & 8 & 34 \\
H3-Max-Turbo & $864\times 496$ & 24 & 10.042 & 8 & 28 \\
Gemini-Omni-Flash & $640\times 360$ & 24 & 10.000 & 8 & 27 \\
Seedance 2.0& $832\times 480$ & 24 & 10.125 & 8 & 28 \\
SANA-WM & $1280\times 704$ & 16 & 10.000 & 8 & 18 \\
LingBot-World 2.0 & $832\times 464$ & 16 & 10.000 & 8 & 18 \\
Matrix-Game 3.5 & $1280\times 704$ & 16 & 10.000 & 8 & 18 \\
Echo-WM-Flash & $1280\times 704$ & 16 & 10.000 & 8 & 18 \\
\bottomrule
\end{tabular}
\vspace{\baselineskip}
\end{table}

\section{Evaluator Validity}
\label{app:evaluator-validity}
We report two complementary checks: agreement between repeated structural-claim judgments and agreement between evaluator decisions and manual review. Table~\ref{tab:evaluator-validity} gives the number of comparisons and exact matches for each check. The repeatability check contains 200 pairs of judgments on the same structural claims. The manual audit contains 150 object--frame decisions for each of association, visibility, identity, and structure, yielding 600 dimension-specific comparisons.

\textbf{Structural-Judge Repeatability.} Structural-claim repeatability concerns the VLM branch of dynamic structure. Agreement requires the same categorical judgment for a claim across the two evaluations: supported, contradicted, or unknown. Exact label agreement measures repeatability. The two calls agree on 181 of 200 claims, giving 90.5\% agreement and 19 disagreements. Repeatability alone cannot establish correctness. A judge can reproduce the same error.

\textbf{Object--Frame Audit and Agreement Metric.} The audit contains 150 decisions for each of four dimensions, using the following rubric.
\begin{itemize}
    \item \textbf{Association:} choose the generated observation corresponding to the reference instance, or label it null or ambiguous. 
    \item \textbf{Visibility:} assign expected-visible, occluded, out-of-view, too-small, or unknown.
    \item \textbf{Identity:} assign preserved, changed, or unassessable using input-specific appearance, including color pattern, texture, and distinctive parts.
    \item \textbf{Structure:} assign preserved, changed, or unassessable for input-supported properties. Static objects are judged for shape and part-layout preservation under viewpoint change; dynamic objects allow articulation and deformation consistent with their kinematic class while retaining supported proportions, connectivity, and material continuity.
\end{itemize}
We report exact agreement as $100\times n_{\mathrm{agree}}/N$, where $n_{\mathrm{agree}}$ is the reported number of matching judgments and $N$ is the total number of comparisons. For repeatability, the comparison is between two judgments of the same claim; for the manual audit, it is between the evaluator decision and manual review. Association uses exact assignment agreement, while visibility, identity, and structure use categorical label agreement. Percentages are computed directly from the counts and rounded to one decimal place.

\begin{table}[t]
\centering
\small
\tablecaption{\textbf{Evaluator-validity results.} Repeatability uses 200 claim pairs. Each manual-audit dimension contains 150 decisions. Agreement is the number of agreeing judgments divided by $N$, expressed as a percentage.}
\label{tab:evaluator-validity}
\begin{tabular}{cccc}
\toprule
Check & $N$ & Agreeing & Agreement (\%) \\
\midrule
Structural-claim repeatability & 200 & 181 & 90.5 \\
\midrule
Association: exact assignment & 150 & 139 & 92.7 \\
Visibility: five labels & 150 & 136 & 90.7 \\
Identity: three labels & 150 & 129 & 86.0 \\
Structure: three labels & 150 & 125 & 83.3 \\
\bottomrule
\end{tabular}
\vspace{\baselineskip}
\end{table}

Association, visibility, identity, and structure have 11, 14, 21, and 25 disagreements, respectively. Across these four dimensions, 529 of 600 decisions agree with manual review, giving a descriptive pooled agreement of 88.2\%. This denominator counts dimension-specific decisions rather than independent samples and excludes the separate repeatability check. Structure has the largest number of disagreements in this audit. The 90.5\% repeat agreement measures judge stability rather than correctness against manual review.

\section{Rationale for OPIS scoring.}
\label{app:scoring}
Although object instances are the basic units of evaluation, OPIS aims to assess memory fidelity at the frame and video levels. For a fixed image extent, we argue that the penalty for an object-memory failure should not be diluted merely because the scene contains more correctly preserved instances. We therefore use an absolute failure criterion rather than the proportion of failed instances: a single confirmed failure is sufficient to invalidate the corresponding frame-level component. This design deliberately measures whether all evaluable instances are preserved together, rather than the average success rate of an individual instance. Accordingly, lower scores in denser scenes indicate greater difficulty in preserving the complete observable inventory, but do not by themselves establish a decline in per-instance memory fidelity.

\section{Case Study}
\label{app:case-study}
\begin{figure*}[p]
\centering
\includegraphics[width=\textwidth]{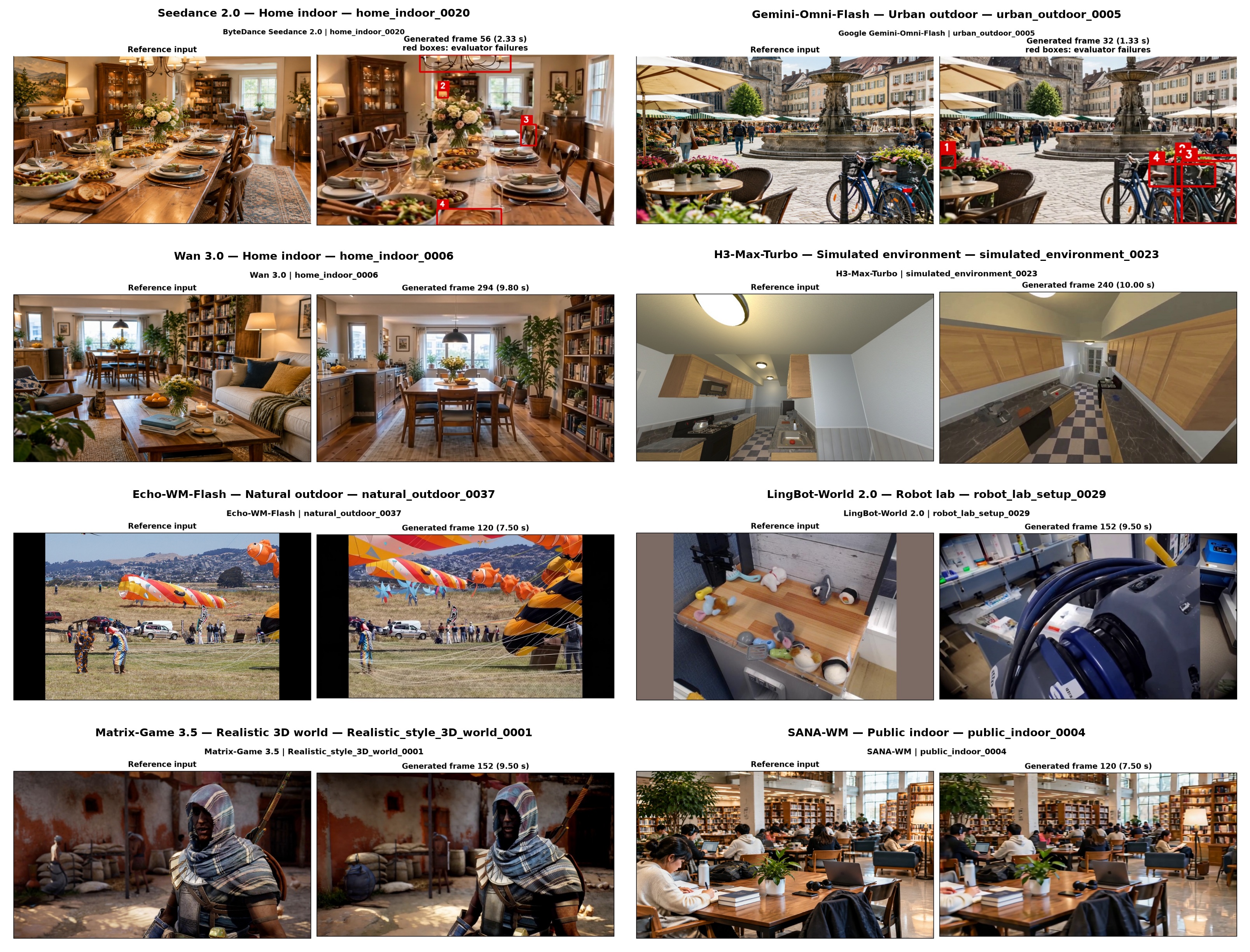}
\caption{Eight visual case studies. Each tile pairs the reference input (left) with a frame from that model's evaluated generation (right). The selected examples span seven subcategories and show visible instance-appearance or structural drift. Red boxes in the two archived panels indicate the evaluator observations used in the original comparison.}
\label{fig:eval-failure-overview}
\end{figure*}

Figure~\ref{fig:eval-failure-overview} shows one example for each evaluated model, spanning seven dataset subcategories. We manually compared every displayed frame with its reference and retained only cases with a visible preservation failure; cases whose low score was attributable only to an incorrect evaluator association were excluded. Table~\ref{tab:eval-failure-sources} reports the corresponding case-level scores.

\begin{table*}[!ht]
\centering\scriptsize
\tablecaption{Eight visual case studies, one per model, spanning seven dataset subcategories. P, I, S, and OPIS are strict case-level scores on a 0--100 scale.}
\label{tab:eval-failure-sources}
\resizebox{\textwidth}{!}{%
\begin{tabular}{llllrrrr}
\toprule
Model & Subcategory & Released case & Visible change & $P$ & $I$ & $S$ & OPIS \\
\midrule
Seedance 2.0 & Home indoor & \texttt{home\_indoor\_0020} & tabletop and lighting details drift & 100.0 & 2.5 & 52.4 & 41.9 \\
Gemini-Omni-Flash & Urban outdoor & \texttt{urban\_outdoor\_0005} & foreground bicycles and furniture drift & 63.0 & 5.2 & 24.3 & 24.4 \\
Wan 3.0 & Home indoor & \texttt{home\_indoor\_0006} & cat, coffee table, and sofa area disappear & 94.1 & 4.2 & 14.7 & 26.4 \\
H3-Max-Turbo & Simulated environment & \texttt{simulated\_environment\_0023} & cabinetry and fixture layout changes & 92.6 & 27.5 & 24.4 & 39.3 \\
Echo-WM-Flash & Natural outdoor & \texttt{natural\_outdoor\_0037} & inflatable forms multiply and change shape & 100.0 & 11.4 & 51.2 & 45.0 \\
LingBot-World 2.0 & Robot lab & \texttt{robot\_lab\_setup\_0029} & tabletop objects and equipment are replaced & 100.0 & 4.9 & 60.8 & 46.3 \\
Matrix-Game 3.5 & Realistic 3D world & \texttt{Realistic\_style\_3D\_world\_0001} & character and background object details change & 84.6 & 73.6 & 66.2 & 72.8 \\
SANA-WM & Public indoor & \texttt{public\_indoor\_0004} & foreground furniture and people change & 100.0 & 25.4 & 0.0 & 30.2 \\
\bottomrule
\end{tabular}
}
\end{table*}

\begin{figure*}[p]
\centering
\includegraphics[width=\textwidth]{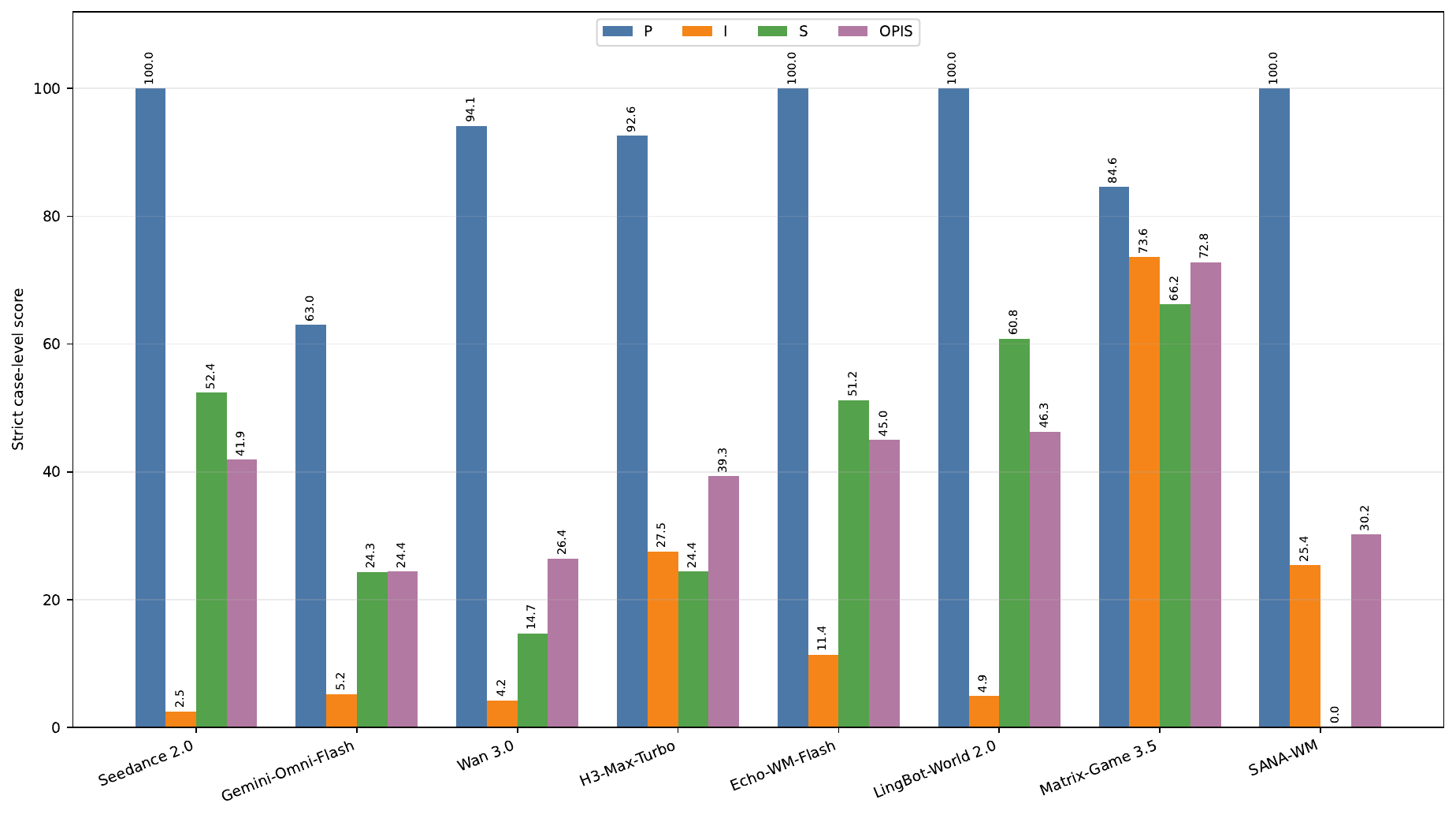}
\caption{Strict case-level Presence (P), Identity (I), Structure (S), and OPIS scores for the eight displayed cases. The chart visualizes the same values reported in Table~\ref{tab:eval-failure-sources}.}
\label{fig:eval-failure-scores}
\end{figure*}

\subsection{Four detailed visual comparisons}

We show four of the eight cases below. The selection includes two image-to-video models and two camera-conditioned world models. Each figure places the reference and generated frame in one row.

\textbf{Seedance 2.0: \texttt{home\_indoor\_0020}.} For Seedance 2.0 on \texttt{home\_indoor\_0020}, the scores are $P=100.0$, $I=2.5$, $S=52.4$, and OPIS $=41.9$. At frame 56 (2.33 seconds), the dining-room composition remains visible, while the chandelier branches, several chairs, and small tabletop objects differ from the reference.

\begin{figure*}[p]
\centering
\includegraphics[width=\textwidth]{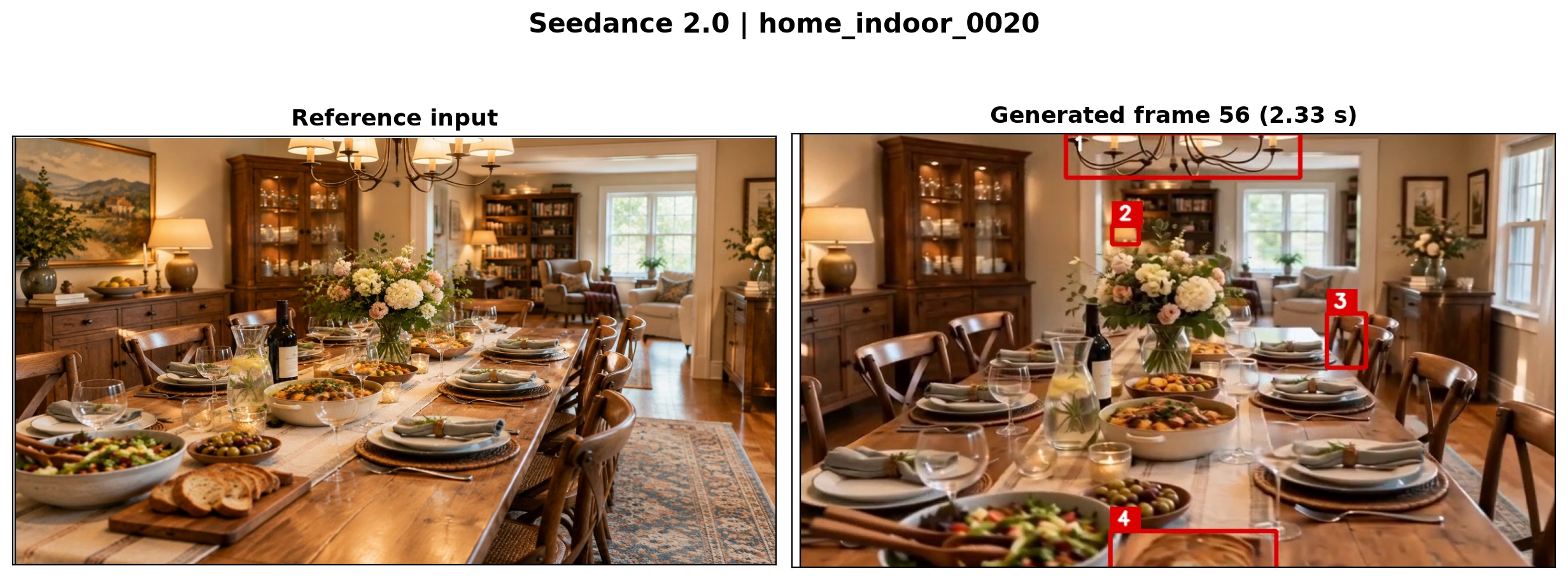}
\caption{Seedance 2.0 on \texttt{home\_indoor\_0020}: full-frame reference and generated-frame comparison.}
\label{fig:case-detail-seedance-full}
\end{figure*}

\textbf{Wan 3.0: \texttt{home\_indoor\_0006}.} For Wan 3.0 on \texttt{home\_indoor\_0006}, the scores are $P=94.1$, $I=4.2$, $S=14.7$, and OPIS $=26.4$. At frame 294 (9.80 seconds), the cat, coffee table, and sofa area visible in the reference are absent, leaving the dining table and bookcase as the dominant foreground structures.

\begin{figure*}[p]
\centering
\includegraphics[width=\textwidth]{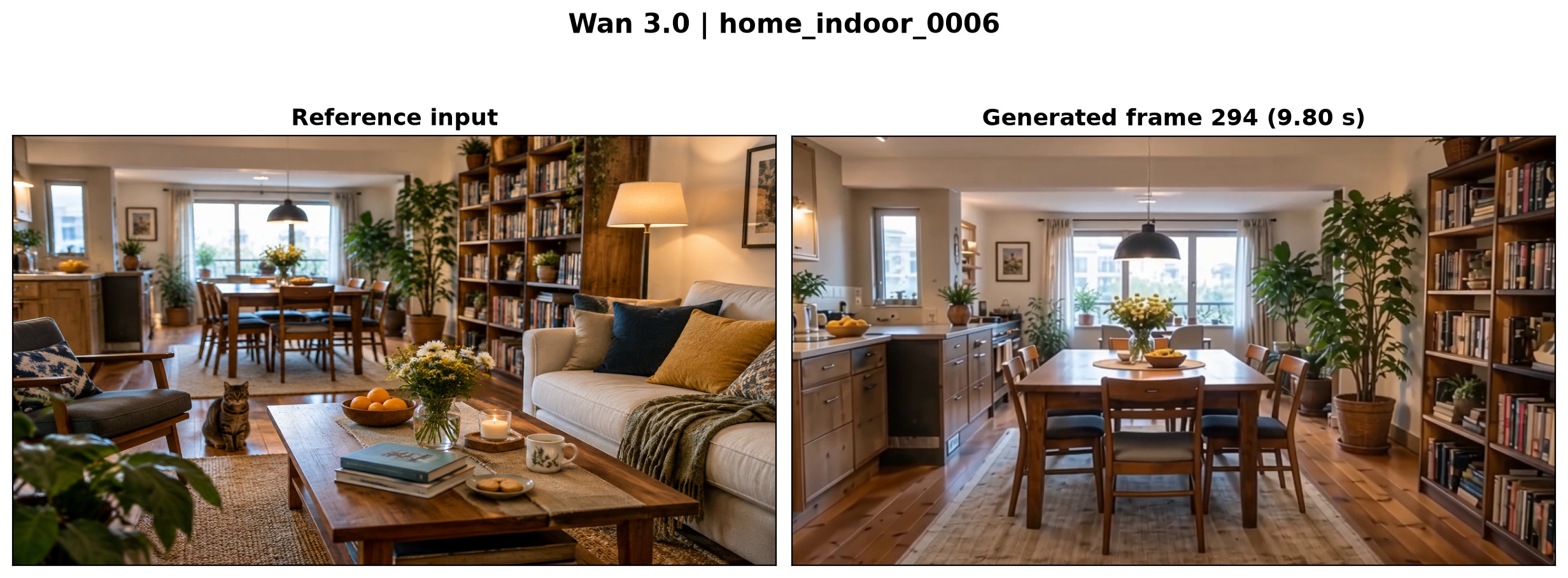}
\caption{Wan 3.0 on \texttt{home\_indoor\_0006}: full-frame reference and generated-frame comparison.}
\label{fig:case-detail-wan-full}
\end{figure*}

\textbf{LingBot-World 2.0: \texttt{robot\_lab\_setup\_0029}.} LingBot-World 2.0 is camera-conditioned. On \texttt{robot\_lab\_setup\_0029}, its scores are $P=100.0$, $I=4.9$, $S=60.8$, and OPIS $=46.3$. At frame 152 (9.50 seconds), after the requested return interval, the original tabletop arrangement is no longer restored: a large dark object fills the foreground and the remaining items differ in color and shape.

\begin{figure*}[p]
\centering
\includegraphics[width=\textwidth]{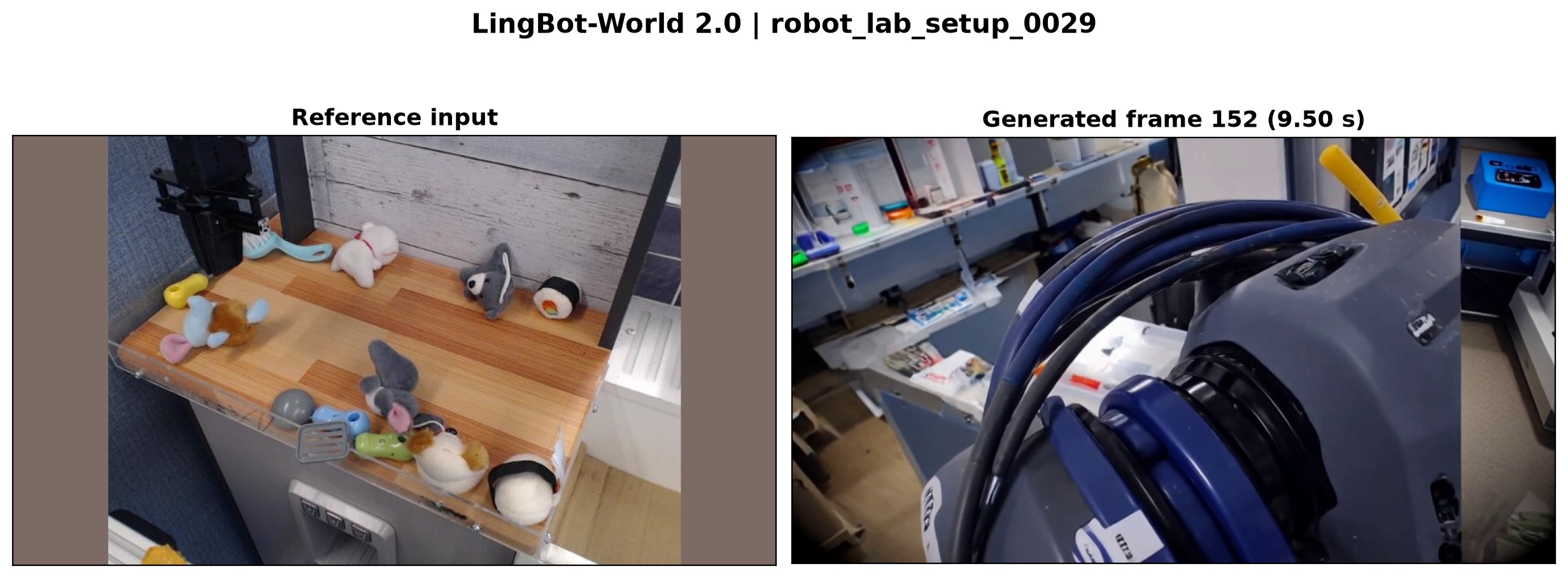}
\caption{LingBot-World 2.0 on \texttt{robot\_lab\_setup\_0029}: full-frame reference and generated-frame comparison.}
\label{fig:case-detail-lingbot-full}
\end{figure*}

\textbf{Matrix-Game 3.5: \texttt{Realistic\_style\_3D\_world\_0001}.} Matrix-Game 3.5 is camera-conditioned. On \texttt{Realistic\_style\_3D\_world\_0001}, its scores are $P=84.6$, $I=73.6$, $S=66.2$, and OPIS $=72.8$. At frame 152 (9.50 seconds), after the requested return interval, the hooded character's facial and armor details differ, while the standing figure and sacks at the left are replaced by a blue bag and a different container arrangement.

\begin{figure*}[p]
\centering
\includegraphics[width=\textwidth]{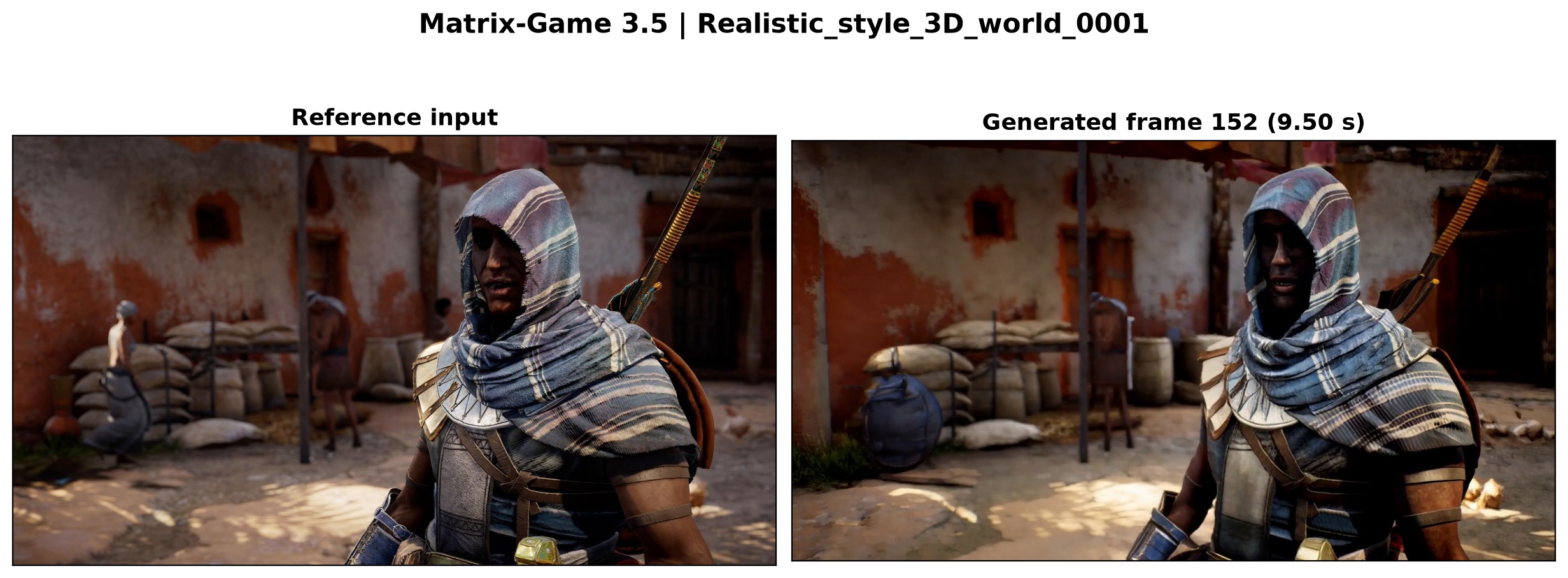}
\caption{Matrix-Game 3.5 on \texttt{Realistic\_style\_3D\_world\_0001}: full-frame reference and generated-frame comparison.}
\label{fig:case-detail-matrix-full}
\end{figure*}

\end{document}